%% file: main.tex
\makeatletter
\@namedef{ver@everyshi.sty}{}
\makeatother
\documentclass[10pt,twocolumn,letterpaper]{article}

\usepackage{iccv}
\usepackage{times}
\usepackage{epsfig}
\usepackage{graphicx}
\usepackage{amsmath}
\usepackage{amssymb}

\usepackage{booktabs}
\usepackage{color}
\usepackage{gensymb}
\usepackage{tabularx}
\usepackage{tikz}
\usepackage[utf8]{inputenc}

\usepackage{enumitem}
\setitemize{noitemsep,topsep=0pt,parsep=0pt,partopsep=0pt}

\usepackage[pagebackref=true,breaklinks=true,colorlinks,bookmarks=false]{hyperref}

\iccvfinalcopy 

\def\iccvPaperID{3757} 

\ificcvfinal\fi

\input{macros} 

\newcommand{\PAR}[1]{\vskip4pt \noindent {\bf #1~}}

\begin{document}

\title{(Just) A Spoonful of Refinements Helps the Registration Error Go Down}

\author{Sérgio Agostinho$^1$
\quad
Aljoša Ošep$^2$
\quad
Alessio {Del Bue}$^3$
\quad
Laura Leal-Taixé$^2$\\
$^1$ Instituto Superior Técnico,  Portugal \quad $^2$Technical University of Munich, Germany\\ $^3$Fondazione Istituto Italiano di Tecnologia, Italy\\
$^1${\tt\small sergio.agostinho@tecnico.ulisboa.pt} \quad
$^2${\tt\small \{aljosa.osep, leal.taixe\}@tum.de}\\
$^3${\tt\small alessio.delbue@iit.it}
}

\maketitle
\ificcvfinal\thispagestyle{empty}\fi

\begin{abstract}
    We tackle data-driven 3D point cloud registration. Given point correspondences, the standard Kabsch algorithm provides an optimal rotation estimate. This allows to train registration models in an end-to-end manner by differentiating the SVD operation. However, given the initial rotation estimate supplied by Kabsch, we show we can improve point correspondence learning during model training by extending the original optimization problem. In particular, we linearize the governing constraints of the rotation matrix and solve the resulting linear system of equations. We then iteratively produce new solutions by updating the initial estimate.  
    Our experiments show that, by plugging our differentiable layer to existing learning-based registration methods, we improve the correspondence matching quality. This yields up to a 7\% decrease in rotation error for correspondence-based data-driven registration methods. 
\end{abstract}


\input{01_introduction_arxiv}

\input{02_related_work_arxiv}

\input{03_method_arxiv}

\input{04_experiments_arxiv}

\section{Conclusion}
In this paper, we presented a differentiable rotation estimation approach that can be used in combination with the Kabsch algorithm. 
We show that it improves correspondence matching quality resulting in improved registration.  
We expect our method to benefit future methods tackling learning-based end-to-end correspondence-based registration method. %

\PAR{Acknowledgements.}The authors would like to thank
Franziska Gerken, Mark Weber,
and Qunjie Zhou
for their insightful suggestions.
This work was partly funded by FCT through grant PD/BD/114432/2016 and project UIDB/50009/2020. It was also supported by project 24534 - INFANTE, funded by the COMPETE 2020 and Lisboa 2020 programs, under the PORTUGAL 2020 Partnership Agreement, through the European Regional Development Fund.

{\small
\bibliographystyle{ieee_fullname}
\bibliography{main}
}

\clearpage
\input{05_supplementary_arxiv}

\end{document}

%% file: macros.tex
\def\vec#1{\mathchoice%
	{\mbox{\boldmath $\displaystyle\bf#1$}}
	{\mbox{\boldmath $\textstyle\bf#1$}}
	{\mbox{\boldmath $\scriptstyle\bf#1$}}
	{\mbox{\boldmath $\scriptscriptstyle\bf#1$}}}

\def\mat#1{\mathchoice{\mbox{\boldmath$\displaystyle\tt#1$}}
	{\mbox{\boldmath$\textstyle\tt#1$}}
	{\mbox{\boldmath$\scriptstyle\tt#1$}}
	{\mbox{\boldmath$\scriptscriptstyle\tt#1$}}}

\DeclareMathOperator{\veco}{vec}
\DeclareMathOperator{\tr}{tr}

\DeclareMathOperator*{\argmin}{arg\,min}
\DeclareMathOperator*{\st}{s.t.}
\newcommand{\R}{\mathbb{R}}

\newcommand{\cbox}[1]{\tikz[baseline=-0.5ex]\draw[#1, line width=3, ](0,0) -- (0.2, 0);}

\definecolor{csti}{RGB}{108, 142, 191} 
\definecolor{cstii}{RGB}{214, 182, 86} 
\definecolor{cstiii}{RGB}{150, 115, 166} 

%% file: 01_introduction_arxiv.tex
\section{Introduction}

Finding a geometrical transformation that aligns two point sets is at the core of several down-stream tasks such as range data fusion~\cite{Izadi2011ACMUIST}, ego- or object pose tracking~\cite{Gross193DV,Held14RSS,Zhang2019ICCV}, 3D shape completion~\cite{gu20eccv} and camera re-localization~\cite{barsan18corl}. 
This challenging problem has a long-standing research history~\cite{gower75Psychometrika,Umeyama1991CAL,Besl92PAMI,Tsin04ECCV,Myronenko2010TPAMI}, as the correspondence between point clouds is usually not known, or it may not even explicitly exist. Other challenges include the fact that 3D sensors often observe object or scene surfaces only partially, that the density of scans varies with respect to the distance from the sensor, and that point clouds are usually corrupted by severe noise and outliers. 

\begin{figure}[t]
\begin{center}
    \includegraphics[width=1.0\linewidth]{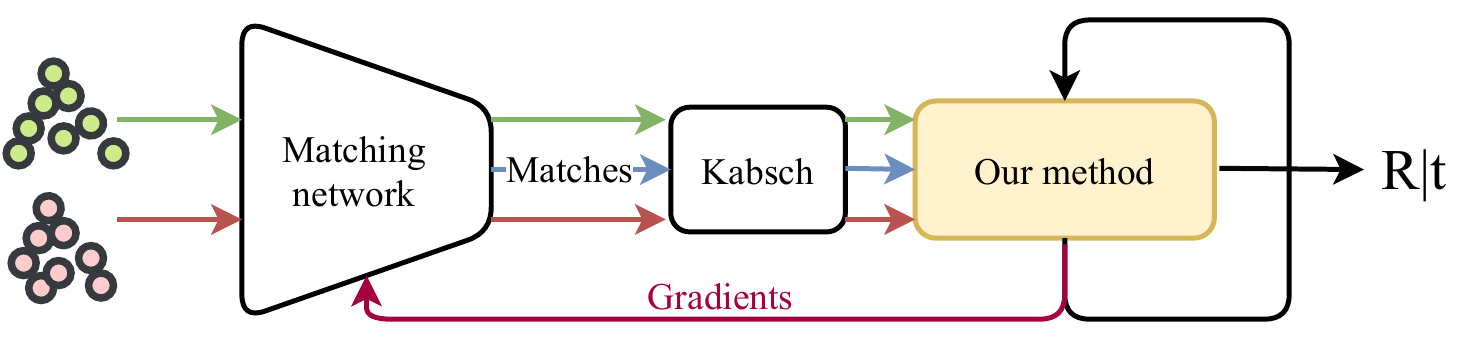}
\end{center}
\caption{We propose a novel, differentiable rotation estimator. Given a trainable, correspondence-based registration method and initial rotation estimate (by, \eg, Kabsch algorithm), our iterative rotation estimator performs a series of pose refinements to produce gradients that guide the matching network towards improving point correspondences.
}
\label{fig:teaser}
\end{figure}

Existing optimization-based methods typically solve point cloud registration by finding a transformation that minimizes the distance between two point sets, according to some criterion, \eg, Chamfer distance~\cite{Fitzgibbon2003IVC}. Their performance critically depends on the quality of the 3D point correspondences between the point sets~\cite{johnson99pami, salti14cviu, rusu09icra, Khoury17ICCV, Zeng17CVPR, yew18eccv, Deng18CVPR}. 
To cope with the limitations of prior heuristic approaches, recent methods~\cite{Zeng17CVPR,yew18eccv,Deng18CVPR,deng18eccv,Choy2019ICCV,Gross193DV,Hertz2020CVPR} leverage the representational power of deep neural networks to learn an estimate of the transformation that aligns point clouds in an end-to-end manner.

To ensure that the estimated matrix is a valid rigid transformation matrix, state-of-the-art learning-based methods~\cite{wang19ICCV,yew20CVPR,choy20CVPR} employ the Kabsch algorithm~\cite{gower75Psychometrika} and thus need to differentiate through the SVD operation~\cite{papadopoulo00ECCV}.
Kabsch produces globally optimal estimates for a given set of correspondences. However, if the point matches are not perfect, we can design other pose oriented geometric incentives to guide the correspondence network to learn better matches. 

To this end, we formulate a local approximation of the correspondence distance minimization problem and  show 
that this helps the network to produce better correspondences.
Our method takes the estimate produced by Kabsch  as input and it performs a number of recurrent iterative steps that over time push the matching network to improve correspondence matching quality (see~\autoref{fig:teaser}).
As rotation matrices are governed by non-linear constraints, we propose a linearization of these constraints around the computed initial estimate and solve the optimization problem using the method of Lagrange multipliers. The result is a linear system of equations, where the new rotation estimate can be extracted in closed-form. This estimate is recurrently refined over a fixed number of iterations, akin to the classic approach from Drummond and Cipolla~\cite{Drummond2002PAMI}.
In comparison to Kabsch, the use of linearized constraints makes our estimator increasingly sensitive to particular geometric configurations of point clouds, that result in estimates that might diverge from the original Kabsch estimates.
We discuss these configurations in more detail in the supplementary material.
However, Kabsch also benefits from a network that is encouraged to prevent these particular configurations.

We show experimentally that our approach is an add-on module that can be used in combination with different learning-based
methods for point cloud registration, such as Deep Closest Point~\cite{wang19ICCV} and RPM-Net~\cite{yew20CVPR}. 
We improve on the results of two state-of-the-art methods, in a registration task on the ModelNet40~\cite{Wu15CVPR} dataset. 
Moreover, we show that when using our layer, there is no need to impose an additional loss that aids the matching network (as used in, \eg,~RPM-Net~\cite{yew20CVPR}), as the correspondences improve during the training implicitly. 
In summary, our contributions are the following:
i) We propose a novel, differentiable rotation estimation layer that promotes the trainable matching network to improve correspondence matching quality;
ii) We show that this differentiable rotation estimator can be obtained by linearizing the governing constraints for rotation matrices around the initial solution in an iterative fashion, resulting in linear constrained optimization problem with closed-form solution;
iii) We augment two state-of-the-art deep global registration methods with our layer and improve upon them in the task of point cloud registration on the ModelNet40 benchmark.
Our module can improve learning-based registration methods,
at the minimal cost of adding a parameter-free layer during training.\footnote{Our implementation and trained models can be found at \url{https://github.com/SergioRAgostinho/just-a-spoonful}}

%% file: 02_related_work_arxiv.tex
\begin{figure*}[t]
\begin{center}
    \includegraphics[width=0.9\linewidth]{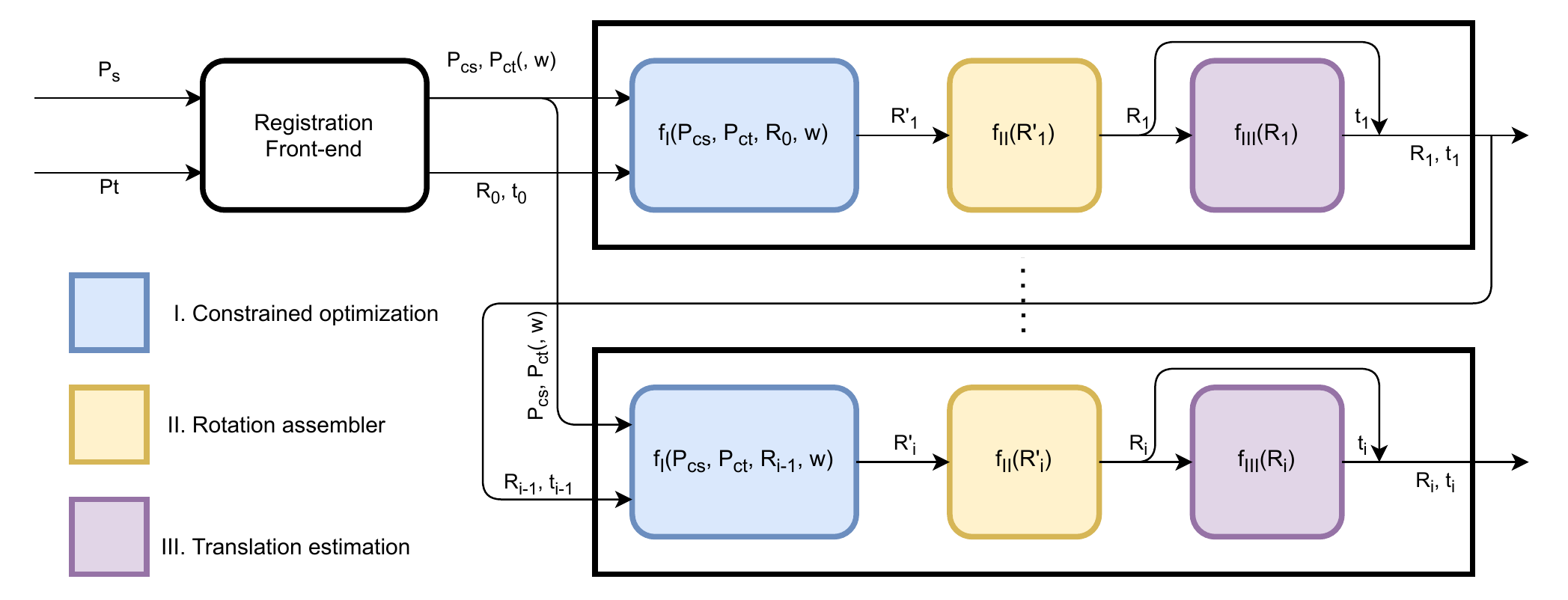}
\end{center}
\caption{An overview of our proposed method: The variables $\mat P_s$, $\mat P_t$ represent the source and target point clouds, and $\vec{w}$ is a set of optional weights pondering each correspondence. The trainable \textit{matching network} produces 3D point correspondences $\mat P_{cs}$ and $\mat P_{ct}$ from both point clouds, an initial pose estimate $(\mat R_0, \vec t_0)$ and an optional, represented in parenthesis $(\cdot)$, vector of weights $\vec w \in \R^N_+$ ranking each correspondence.
At refinement iteration $i$, we first produce an approximate rotation estimate $\mat R'_i$ which minimizes the point-to-point distance between correspondences \textit{(constrained optimization module)}. Then, our \textit{rotation assembler} 
generates a proper rotation matrix from the approximate rotation estimate $\mat R'_i$. Finally, we estimate the optimal translation $\vec t_i$ between correspondences using \textit{translation estimator}, 
given the rotation estimate from the \textit{rotation assembler}.
}
\label{fig:overview}
\end{figure*}

%
\section{Related Work}

Point cloud registration is an extensively studied topic, with a considerable literature spanning decades~\cite{gower75Psychometrika,Umeyama1991CAL,Besl92PAMI,Tsin04ECCV,Myronenko2010TPAMI}. In this section, we present some of the most relevant optimization-based methods and recent data-driven approaches.

\PAR{Optimization-based methods.}
Consolidated solutions for point clouds registration are based on Iterative Closest Point (ICP)~\cite{chen92ivc, Besl92PAMI} and its variants~\cite{rusinkiewicz01TDIM, Segal09RSS, Bouaziz13CGF}. These methods alternate between the correspondence search and optimal pose estimation given such matches in the point sets.
Standard ICP approaches use a nearest neighbor strategy in coordinate space to establish correspondences. This approach is sensitive to the initial transformation estimate and is therefore considered to be a local method. 
With exact correspondences, estimating the optimal pose can be formulated as an Orthogonal Procrustes problem, that can be minimized using the Kabsch algorithm~\cite{gower75Psychometrika}. 
For this reason, the community invested efforts in developing robust point descriptors for finding the best correspondences. Most of them  rely on  hand-crafted descriptors, \eg, SPIN~\cite{johnson99pami}, SHOT~\cite{salti14cviu} and FPFH~\cite{rusu09icra}. 
Instead, recent efforts privilege a data-driven approach by leveraging point cloud encoders~\cite{Qi17CVPRb,Li2018NeuRIPS} to learn point descriptors~\cite{Khoury17ICCV, Deng18CVPR, Choy2019ICCV}. 

Beyond improving feature detection and matching, global registration methods often adopt a robust cost function to improve robustness to outliers~\cite{Zhou16ECCV}. The work of~\cite{Yang2020arXiv} explicitly focuses on identifying outliers in the point matches, while \cite{Yang19ICCV} finds a globally optimal solution through branch-and-bound systematic search in the SE(3) solution space. Different from previous approaches, \cite{Le2019CVPR}  uses sophisticated sampling and graph matching mechanisms to bypass having to establish putative correspondences.

\PAR{Data-driven methods.}
Recent advances in the area of point cloud representation learning~\cite{Qi17CVPR} paved the way towards data-driven methods for point cloud alignment~\cite{Gross193DV,Goforth19CVPR,wang19ICCV,choy20CVPR, Hertz2020CVPR,deng19CVPR}. 
While it is possible to use a learned feature detector~\cite{Khoury17ICCV, deng18eccv, Zeng17CVPR, yew18eccv} in combination with global registration methods~\cite{Zhou16ECCV}, the ultimate goal is to optimize the point representation with respect to the final task in an end-to-end manner.

Several methods directly learn to regress the transformation between two point clouds~\cite{Goforth19CVPR, Gross193DV, Pais2020CVPR, Hertz2020CVPR}.
PoinNetLK~\cite{Goforth19CVPR} requires an initial transformation estimate and proceeds iteratively by minimizing the distances between the point cloud embeddings. AlignNet~\cite{Gross193DV} computes relative transformation in a single shot by first estimating the canonical pose, followed by the estimation of residual transformation. 
PointGMM~\cite{Hertz2020CVPR} leverages hierarchical Gaussian Mixture Models to learn a multi-scale representation of the point cloud that disentangles orientation and shape in the embedding space. The relative transformation can then be computed by estimating a canonical pose of each point cloud.  

Other methods explicitly establish correspondences. 
3DRegNet~\cite{Pais2020CVPR} leverages a correspondence classification mechanism inspired by Kim~\etal~\cite{moo18cvpr}  and regresses the rigid transformation by optimizing directly in the SE(3) manifold.
Other methods use Kabsch~\cite{gower75Psychometrika} to ensure that the estimated transformation is a valid euclidean transformation, composed by a proper rotation matrix. 
Deep Closest Point (DCP)~\cite{wang19ICCV} employs a pointer network module~\cite{Vinyals2015NeurIPS} to establish soft correspondences between two point clouds based on the learned embeddings. 
Deep Global Registration~\cite{choy20CVPR} additionally employs a network module~\cite{Choy2020bCVPR} for correspondence confidence weighting, used in combination with a weighted variant of Kabsch. 
RPM-Net~\cite{yew20CVPR} couples local and global spatial coordinates together with handcrafted point-pair features~\cite{rusu09icra}, as input to a PPFNet feature encoder~\cite{Deng18CVPR}. 
We show experimentally that our method is a worthy add-on to be used in combination with end-to-end trainable correspondence-based methods, such as Deep Closest Point~\cite{wang19ICCV} and RPM-Net~\cite{yew20CVPR}, to improve the registration performance.

%% file: 03_method_arxiv.tex
%

\section{Method}
\label{sec:method}

In this section, we detail our differentiable iterative refinement method that can be added as a complementary step to any correspondence-based registration deep learning method, as can be seen in~\autoref{fig:overview}. 
Inputs to our method are an initial rotation estimate, \eg, supplied by Kabsch, 
a set of point correspondences estimated by any trainable matching network and, optionally, a set of weights ranking the quality of these correspondences. Note, these correspondences are not necessarily correct, in fact, we will show we improve them during the model training thanks to our method. 
We then perform the following steps iteratively (\autoref{fig:overview}): 

\PAR{Constrained optimization} (\cbox{csti}): We produce an approximate rotation estimate by linearizing governing constraints for rotation matrices around the previous estimate. 
This module minimizes the weighted point-to-point distance between correspondences, however, the resulting matrix is not necessarily a valid rotation matrix (see~\autoref{sec:refinements}). 
\PAR{Rotation assembler} (\cbox{cstii}): We convert the matrix of the previous step into a valid rotation matrix by applying Gram-Schmidt orthogonalization to the first two columns of the input and a cross product to generate the final column.

\PAR{Translation estimation} (\cbox{cstiii}): Given an input rotation we can compute the optimal translation vector in a closed-form (see~\autoref{sec:preliminaries}).

\medskip

These operations define our novel layer, which refines the rotation estimates iteratively. However, since the pose returned from Kabsch is already optimal, our main contribution is not in improving the final pose directly but instead conditioning the matching part of the network towards learning better correspondences. 
Empirically we show that our novel differentiable rotation estimator aids correspondence matching registration even though we only impose supervision on the pose.

\subsection{Preliminaries}
\label{sec:preliminaries}
Given a set of correspondences between the source and target point clouds $\mat P_s, \mat P_t \in \R^{N\times 3}$ (see~\autoref{fig:overview}) our aim is to find a rigid transformation ($\mat R$, $\vec t$) that minimizes the following error:
\begin{subequations}
\label{eq:original}
\begin{eqnarray}
\argmin_{\mat R, \vec t} & \sum_{i=1}^N w_i \|\vec{p_t}_i  - \mat R \vec{p_s}_i - \vec t\|^2 \label{eq:original-cost}\\
\st & \mat R \in SO(3),
\end{eqnarray}
\end{subequations}
where $\vec w \in \R_+^N$ represents the (optionally supplied) set of weights and $\vec{p_s}_i$  and $\vec{p_t}_i$ are individual points of the source and target point clouds.
Given a rotation matrix $\mat R$, we can simply extract the optimal translation vector $\vec t$ in a closed-form as:
\begin{equation}
    \vec t^* = \frac{\sum_{i=1}^N w_i(\vec{p_t}_i  - \mat R \vec{p_s}_i)}{\sum_{i=1}^N w_i} = \bar{\vec{p_t}} - \mat R \bar{\vec{p_s}}, \label{eq:optimal-t}
\end{equation}
where $\bar{\vec{p_t}}$ and $\bar{\vec{p_s}}$ represent the weighted means of the points for each point cloud. 
We further define $\tilde{\vec{p_t}}_i$ and $\tilde{\vec{p_s}}_i$ as the mean-subtracted versions of $\vec{p_t}_i$ and $\vec{p_s}_i$, such that $\tilde{\vec{p_s}} = \vec{p_s} - \bar{\vec{p_s}}$ and $\tilde{\vec{p_t}} = \vec{p_t} - \bar{\vec{p_t}}$. We can then factor out the translation component and Eq. \eqref{eq:original} can be formulated entirely with the respect to the rotation. Back-substituting Eq. \eqref{eq:optimal-t} yields the following simplification:
\begin{subequations}
\label{eq:simplified}
\begin{eqnarray}
\argmin_{\mat R} & \sum_{i=1}^N w_i \|\tilde{\vec{p_t}}_i  - \mat R \tilde{\vec{p_s}}_i\|^2 \label{eq:simplified-cost}\\
\st & \mat R \in SO(3) \label{eq:simplified-constraints}.
\end{eqnarray}
\end{subequations}
The Kabsch algorithm~\cite{gower75Psychometrika} provides a closed-form,
globally optimal solution (given correspondences) to this problem via SVD as follows:
\begin{align}
    \mat H &= \sum_{i = 1}^N w_i\tilde{\vec{p_t}}_i \tilde{\vec{p_s}}^\top_i  \label{eq:kabsch-input}\\
    \mat U, \mat S, \mat V &= \operatorname{svd}(\mat H) \\
    \mat R &= \mat U \operatorname{diag} ([1, 1, \det(\mat U \mat V^\top)])\mat V^\top.
\end{align}
The operator $\operatorname{diag(\:)}$ produces diagonal square matrices, in which the input vector represents the diagonal.
Kabsch can only provide the \emph{correct} rotation estimate for pose estimation if correspondences are also correct. 
Our formulation of the optimization problem (Eq.~\ref{eq:simplified-cost}) as an iterative procedure helps the network to produce better correspondences, as we will show in the experimental section. 

\subsection{Just a spoonful of refinements}
\label{sec:refinements}
We first present the governing constraints of the rotation matrix and then discuss our proposed relaxation to their local linear approximation. 
The membership of $\mat R$ in $SO(3)$ can be expressed as: 
\begin{subequations}
\label{eq:constraints}
\begin{eqnarray}
\mat R^\top \mat R &=& \mat I_3 \label{eq:orthogonality}\\
\det \mat R &=& 1, \label{eq:const-det}
\end{eqnarray}
\end{subequations}
where $\mat I_3$ is the $3\times3$ identity matrix.
These are quadratic and cubic equality constraints, respectively. Eq. \eqref{eq:orthogonality} supplies six constraints and Eq. \eqref{eq:const-det} an additional one.
However, as shown in the supplementary material, given that all expressions are linearized around a point that represents a rotation matrix, the determinant constraint is not longer linearly independent \wrt the orthogonality ones and is therefore redundant. The next stage of the formulation requires that all constraints are linearly independent, so we choose to drop the determinant constrained as it allows to proceed with the formulation taking solely into account the orthogonaly related expressions.

\PAR{Linearization of constraints (\cbox{csti}).}
Denoting our prior rotation estimate with $\mat R_{t-1}$,
we start by linearizing the linearly independent components of Eq.~\eqref{eq:orthogonality} around the initialization $\mat R_{t-1}$,  by only taking into consideration the upper triangle section of the constraints matrix.
We define matrix $c(\mat R) = \mat R^\top \mat R - \mat I_3$ such that $c(\mat R): \R^{3\times3} \to \R^{3\times3}$ and refer to $c_{ij}(\mat R)$ as the element in the $i$-th row and $j$-th column, defined as $c_{ij}(\mat R) = \vec e_i^\top (\mat R^\top \mat R - \mat I_3) \vec e_j$. 
The variables $\vec e_i, \vec e_j \in \R^3$ are Euclidean bases, vectors of zeros with a single element equal to one at the $i$-th and $j$-th elements, respectively. 
Using Taylor expansion around the initial estimate $\mat R_{t-1}$ and retaining only terms up to the first order, leads to the following linearized constraints:
\begin{eqnarray}
c_{ij}^{(1)}(\mat R, \mat R_{t-1}) &= c_{ij}(\mat R_{t-1}) + \tr\left(\mat E^\mathbb{S}_{ij} \mat R_{t-1}^\top (\mat R -  \mat R_{t-1})\right) \label{eq:constraint-orthogonal-linear}\\
\text{for} & i = 1, \dots, 3; \, j=i, \dots, 3, \nonumber
\end{eqnarray}
where $c_{ij}^{(1)}$ is the first-order Taylor approximation, of the $i$-th row and $j$-th column of the orthogonality constraints $c(\mat R)$. 
The matrix 
$\mat E^\mathbb{S}_{ij} = \vec e_i \vec e^\top_j + \vec e_j \vec e^\top_i = \mat E_{ij} + \mat E_{ji} \in \mathbb{S}^3$,
with $\mathbb{S}^3$ representing the space of real symmetric matrices of size $3\times3$.
Full derivations for this approximation are provided in the supplementary material.

\PAR{Langrangian formulation (\cbox{csti}).}
After the relaxation and linearization of our constraints, we now have an optimization problem with a quadratic cost function and linear constraints. Then, we can enforce the (linearized) equality  constraints in Eq.~\ref{eq:constraint-orthogonal-linear} using the method of Lagrange multipliers. Thus, we obtain a closed-form solution that can be formulated as a linear system of equations. We write the new Lagrangian of \eqref{eq:simplified-cost} as: 
\begin{equation}
\mathcal{L}(\mat R, \vec \lambda) = \sum_{i=1}^N \frac{w_i}{2} \|\tilde{\vec{p_t}}_i  - \mat R \tilde{\vec{p_s}}_i\|^2 + \sum_{k=1}^6 \lambda_k c^{(1)}_k(\mat R, \mat R_{t-1}), \label{eq:lagrange}
\end{equation}
where indices $ij$ previously used to specify the row and column of the constraints $c_{ij}^{(1)}(\mat R, \mat R_{t-1})$, are now replaced by the single index $k$, iterating over the upper triangular part of the matrix. The variables $\lambda_k$ represent the Lagrange multipliers of the constraints.
We form a linear system for which the solution returns our newly refined estimate, by computing the gradient of Eq. \eqref{eq:lagrange} with respect to both $\mat R$ and $\vec \lambda$, and setting it to 0. The optimal $\mat R$ and $\vec \lambda$ are determined by solving the linear system: 
\begin{equation}
\begin{bmatrix}
\mat A & \mat B \\ \mat B^\top & 0 \end{bmatrix} \begin{bmatrix}
\veco(\mat R) \\ \vec \lambda
\end{bmatrix} = \begin{bmatrix}
\veco\left(\sum_{i=1}^N w_i \tilde{\vec{p_t}}_i \tilde{\vec{p_s}}^\top_i\right) \\
\vec d
\end{bmatrix},
\end{equation} 
where the operator $\veco$ represents a column-wise vectorization of its input matrix and the matrices $\mat A$ and $\mat B$ are defined as in the following. 
The matrix $\mat A$ is given by $\mat A = \left[ \begin{array}{ccc} \vec{a}_1, & \ldots,  & \vec{a}_9
\end{array} \right]^\top \in \R^{9\times9}$, with each vector  $\vec a_r \in \R^9, \: r = 1,\ldots,9$ such that
\begin{equation}
\vec a_r = \veco\left(\mat E_{mn} (\sum_{i=1}^N w_i \tilde{\vec{p_s}}_i \tilde{\vec{p_s}}^\top_i)\right).
\end{equation}
Matrix $\mat E_{mn} \in \R^{3\times3}$ has all elements equal to 0, except the one in row $m$ and column $n$, which is 1.
Matrix $\mat A$ is constructed from $\veco\left(\frac{\partial \mathcal{L}}{\partial \mat R} = 0\right)$ and retaining all terms that depend on $\mat R$.
Matrix $\mat B = \left[ \begin{array}{ccc} \vec{b}_1, & \ldots,  & \vec{b}_6
\end{array} \right] \in \R^{9\times6}$ is composed by the following columns $\vec b_k$ with $k = 1,\dots, 6$, such that: 
\begin{equation}
\vec b_k = \veco(\mat R_{t-1} \mat E^\mathbb{S}_k),
\end{equation}
where the vector $\vec d \in \R^6$, with each element given by
\begin{equation}
d_k = \tr(\mat E^\mathbb{S}_k) - c_k(\mat R_{t-1}).
\end{equation}
In the supplementary material we provide a detailed derivation on how to compose the linear system of equations, specifically matrices $\mat A$, $\mat B$ and vector $\vec d$.
After solving the linear system of equations, we obtain a newly refined rotation estimate.

\PAR{Producing a rotation matrix from a candidate refinement (\cbox{cstii}).}
Due to the linearization of the original rotation constraints, there is no guarantee that our solution from the optimization module (\autoref{fig:overview}, \cbox{csti}) is a valid rotation matrix. 
One solution to this problem would be to project the matrix to the closest orthogonal matrix using SVD by minimizing the Frobenius norm distance to the input. 
This projection step might be non-differentiable because the gradient is not defined 
if the input matrix is already a valid rotation. %
This problem is discussed in Ionescu~\etal~\cite{Ionescu15ICCV}, where they show that the gradient of an SVD is not defined in the case of the input matrix having equal singular values, as a rotation matrix does. 
Moreover, the closer the singular values are to each other, the more numerically ill-conditioned the gradient becomes. Since we intentionally target having matrices very close to true rotations \ie, having all singular values equal to 1, using
SVD is not an option. 
To generate a new rotation representation from our estimate, we instead adopt the strategy proposed by Zhou~\etal~\cite{Zhou19CVPR} that has close ties to Gram-Schmidt orthogonalization. 
Assume that $\mat R'$ is a $3\times3$ input matrix formed by the columns:
\begin{equation}
    \mat R' = \begin{bmatrix}\vec r'_1 & \vec r'_2 & \vec r'_3 \end{bmatrix}.
\end{equation}
The output matrix is going to be composed by the following three columns:
\begin{eqnarray}
    \vec r_1 &=& \frac{\vec r'_1}{\|\vec r'_1\|}, \label{eq:rot-composition-r1}\\
    \vec r_2 &=& \frac{(\mat I - \vec r_1 \vec r_1^\top)\vec r'_2}{\|(\mat I - \vec r_1 \vec r_1^\top)\vec r'_2\|}, \label{eq:rot-composition-r2}\\
    \vec r_3 &=& \vec r_1 \times \vec r_2. \label{eq:rot-composition-r3}
\end{eqnarray}
These operations are performed in the differentiable orthogonalization step (\autoref{fig:overview}, \cbox{cstii}). 
Contrary to SVD projection, with this method we are not operating close to a gradient singularity. Equations \eqref{eq:rot-composition-r1} and \eqref{eq:rot-composition-r2} have singularities if the denominator is 0. However, as shown in the supplementary material, we are not operating close to this region, leading to a stable training procedure.

\subsection{Augmenting the Loss}

Using our refinement module, we produce $t=1,\ldots,N_r$ additional pose estimates. We apply the loss term to every pose estimate, both initialization and refinements. 
This is applicable to every term that depends on the predicted pose.  
As an example, the original loss function used to train Deep Closest Point~\cite{wang19ICCV} is formulated as 
\begin{equation}
\text{Loss} = \|\mat R^\top \mat R_{gt} - \mat I_3\|^2 + \|\vec t - t_{gt}\|^2 +  \lambda \|\theta\|^2.
\end{equation}
In this expression, $\theta$ represents the networks parameters and  $\lambda$ is a hyper-parameter balancing weight-decay during training. With the new pose estimates the loss now becomes: 
\begin{eqnarray}
\text{Loss} &=& \frac{1}{N_r + 1}\sum_{i=1}^{N_r + 1}\|\mat R^\top_{i}\mat R_{gt} - \mat I_3\|^2 , \nonumber\\
            &+& \frac{1}{N_r + 1}\sum_{i=1}^{N_r + 1}\|\vec t_{i} - t_{gt}\|^2 +  \lambda \|\theta\|^2. \label{eq:aug-loss}
\end{eqnarray}
During training, our refiner
is required to output all possible poses, initializing each new refinement of the pose from the previous iteration.
\emph{We stress that our refinement strategy is only used during training}, initializing each new refinement of the pose from the previous iteration. 
At test time, we output the pose produced by Kabsch, which is optimal given correct correspondences.

\subsection{Understanding the Differences}

The estimator we propose solves a very similar problem to Kabsch, minimizing the same correspondence loss, but under a different set of constraints: a linear approximation of the original second-order equality constraints.
A network trained with and without our additional layers will learn a different set of parameters and will have different registration performance. 
The differences in resulting parameters indicate that the gradients used to optimize the network during training are different.
The loss in Eq.~\eqref{eq:aug-loss} ponders an average pose error between all poses produced: if the poses are all equal, the gradient \wrt to one pose or \wrt to all poses is the same.
In the majority of situations, given a rotation estimate from Kabsch, our estimator will replicate this estimate. However, the linearized constraints make our estimator increasingly sensitive to certain geometric configurations of point clouds. Under these configurations, the estimator will produce a pose estimate that will diverge from Kabsch at each iteration. We stress that in our case, divergence comes paired with the positive effect of facilitating the network to avoid said configurations.
The geometric relationship, e.g. distance and angle, between Kabsch's (constrained) solution and the unconstrained one, is one example of aspects that govern the occurrence of the divergent behavior.
We expand on these ideas and provide additional insightful examples in the supplementary material.

%% file: 04_experiments_arxiv.tex
%
\section{Experimental Evaluation}
\label{sec:experimental}

In this section, we show the merit of our method by improving the accuracy of existing matching-based deep global registration methods, Deep Closest Point~\cite{wang19ICCV} and RPM-Net~~\cite{yew20CVPR}. 
We show that our method helps by improving the quality of correspondence matching and, as a result, improves the pose estimates.
We compare the performance of our full pipeline to several baselines. 
We conduct all experiments employing 5 refinements of our method, a choice supported by our ablation studies,
reported in the supplemental material.
Finally, we show the improvements produced by our method are more than a byproduct of a particular initialization, by conducting multiple training sessions with different weight initializations and showing the average pose error is consistent with our prior benchmarks.

\PAR{Datasets.} We conduct our experiments using ModelNet40~\cite{Wu15CVPR} and 3DMatch~\cite{Zeng17CVPR}  based on the experimental setting of~\cite{wang19ICCV,yew20CVPR}. 
ModelNet40 consists of CAD models containing several symmetrical objects which create pose ambiguities. 
3DMatch is a dataset composed of RGB-D scene fragments.
We create rigid transformations by randomly sampling rotations as three Euler angles from the interval $[0\degree, 45\degree]$ and translations in the range of $[-0.5, 0.5]$ along each axis.

\PAR{Metrics.} To compare our method to prior work, we follow their experimental setting and use the same evaluation metrics, including rotation and translation errors, Chamfer distance, and mean point distance. 
We compute the rotation error as:
\begin{align}
    \Delta \mat R &= \mat R^\top \mat R_{gt} \\
    \angle \Delta \mat R_{\text{iso}} &= \arccos \left(\frac{\operatorname{tr}(\Delta \mat R) - 1}{2}\right) \label{eq:rot-error-iso}\\
    (\angle_z, \angle_y, \angle_x)_{\text{ani}} &= f_{\text{Euler}_{z\to y \to x}}(\Delta \mat R), \label{eq:rot-error-ani}
\end{align}
in both an isotropic \eqref{eq:rot-error-iso} and an anisotropic \eqref{eq:rot-error-ani} forms. The matrices $\mat R$ and $\mat R_{gt}$ stand for the rotation estimate and ground-truth, respectively. The operator $\operatorname{tr}(\cdot)$ stands for the trace of a matrix, $f_{\text{Euler}_{z\to y \to x}}$ is a function which decomposes the rotation matrix in intrinsic Euler angles following the axes sequence $z\to y \to x$. The translation error is computed as the $p-norm$: 
\begin{equation}
    \Delta \vec t = \|\vec t - \vec t_{gt}\|_{p = \{1, 2\}}, \label{eq:trans-error-p}
\end{equation}
with $\vec t$ and $\vec t_{gt}$ standing for the translation estimate and ground-truth. The Chamfer distance between two point sets $\mat P_s$ and $\mat P_t$ is given by:
\begin{align}
\operatorname{CD}(\mat P_s, \mat P_t) &= \frac{1}{|\mat P_s|}\sum_{\vec p_{s_i} \in \mat P_s} \underset{\vec p_{t_j} \in \mat P_t}{\min} \|\vec p_{s_i} - \vec p_{t_j}\|_2^2 \\
&+ \frac{1}{|\mat P_t|}\sum_{\vec p_{t_j} \in \mat P_t} \underset{\vec p_{s_i} \in \mat P_s}{\min} \|\vec p_{s_i} - \vec p_{t_j}\|_2^2. \label{eq:chamfer-dist}
\end{align}
Lastly, the mean point distance is computed as:
\begin{equation}
    \Delta \vec p = \frac{1}{N} \sum_{i = 1}^N \|(\mat R - \mat R_{gt}) \vec p_{s_i} + \vec t -  \vec t_{gt} \|_2. \label{eq:point-dist-error}
\end{equation}
For this metric, we pick the source point cloud to compute the distance without loss of generality.

\subsection{Deep Closest Point with ModelNet40 Data}
\label{sec:exp-dcp}

We augment Deep Closest Point (DCP)~\cite{wang19ICCV} with our refinement stage and evaluate the performance of the final network following their evaluation protocol. 
We use 9,843 meshes for training and 2,468 for testing. From each mesh, we uniformly sample 1024 points based on the face area and scale them to be within a unit sphere. 
DCP's reports the anisotropic rotation error from Eq. \eqref{eq:rot-error-ani} and translation error, computing the Mean Squared Error (MSE), the Root Mean Squared Error (RMSE), and Mean Absolute Error (MAE). 
We present and discuss the simpler scenario of alignment for identical point clouds in the supplementary material.

\noindent{\bf Alignment under Gaussian noise.} In \autoref{table:real-gaussian:modelnet40} we report point cloud registration results obtained on instances of CAD models that were held-out during the model training. Furthermore, we add a Gaussian noise $\mathcal{N}(0, 0.01^2)$, clamped at $[-0.05, 0.05]$ during test time using the model trained on noise-free data. 
To ensure that there are no true correspondences, we applied the noise independently to the source point cloud.\footnote{Note that this is different to the experiment conducted in~\cite{wang19ICCV}, where the noise was not added independently.} 
Our refinement strategy significantly improves the rotation estimator error while incurring in negligible worsening of the translation error. This experiment confirms that we improve the generalization to held-out instances even when 1-to-1 correspondences do not exist. We improve the rotation error by 9\% at the marginal cost of translation accuracy of 0.06\%, when normalized by the maximum magnitude of the rotation and translations sampled. This is an experiment in which DCP performs worse because the perfect correspondence assumption is broken during testing. The interesting aspect is that despite never having access to cases during training in which there were no perfect correspondences, applying our layer assists the network to generalize better to this case.
\begin{table}[t] 
\begin{center}
\resizebox{\linewidth}{!}{%
    \begin{tabular}{lccccccccr}
    \toprule
    Model & RMSE($\mat R$)\degree & MAE($\mat R$)\degree  & RMSE($\vec t$) & MAE($\vec t$)\\ 
    \midrule
    DCP-v2 & 12.974750 & 5.830703 & \textbf{0.003941} & \textbf{0.002502} \\
    \textbf{DCP-v2 + ours}  & \textbf{8.938098} & \textbf{4.373459} &  0.004221 & 0.002777 \\
    \bottomrule
    \end{tabular}
}
\end{center}
\caption{Deep Closest Point on ModelNet40: Test on objects with Gaussian noise only added to the source point cloud. The noise is only added at test time, using the models trained under noiseless conditions.
\emph{There are no perfect correspondences between the source and target point clouds.}\label{table:real-gaussian:modelnet40}}

\end{table}

\noindent{\bf Alignment for unseen categories.} In \autoref{table:unseen:modelnet40}, we report results for unseen categories. 
In prior experiments, the network was trained with all categories, with separate instances being part of the training and testing datasets. In this experiment, all instances from the first 20 categories are used for training and all instances of the last 20 categories are used for testing, putting the generalization capabilities to the test.
Our method achieves the lowest errors in all metrics reported. In particular, we improve the rotation error by 2\% and the translation error by 0.1\%, when normalized by the maximum magnitude of the rotation and translations sampled. This is an experiment in which despite the unseen categories, there are still perfect correspondences in place, which is a more favorable scenario for DCP and that is why our improvements are less substantial than when Gaussian noise was added.

\begin{table}[t]
\begin{center}
\resizebox{\linewidth}{!}{%
    \begin{tabular}{lcccccccr}
    \toprule
    Model & RMSE($\mat R$)\degree & MAE($\mat R$)\degree  & RMSE($\vec t$) & MAE($\vec t$)\\ 
    \midrule
    ICP &  29.876431 & 23.626110 & 0.293266 & 0.251916 \\
    Go-ICP \cite{Yang15TPAMI} & 13.865736 & 2.914169 & 0.022154 & 0.006219 \\
    FGR \cite{Zhou16ECCV} & 9.848997 & 1.445460 & 0.013503 & 0.002231 \\
    PointNetLK \cite{Goforth19CVPR} & 17.502113 & 5.280545 & 0.028007 & 0.007203 \\
    \midrule
    DCP-v2 & 3.150191 & 2.007210 & 0.005039 & 0.003703 \\
    \textbf{DCP-v2 + ours}  & \textbf{2.051713} & \textbf{1.431898} & \textbf{0.004543} & \textbf{0.003333} \\
    \bottomrule
    \end{tabular}
}
\end{center}
\caption{Deep Closest Point on ModelNet40: Test on unseen categories.\label{table:unseen:modelnet40}}
\end{table}


\subsection{RPM-Net with ModelNet40 Data}
\label{sec:rpm-net}

Similar to DCP, we augment RPM-Net~\cite{yew20CVPR} with our proposed layer and evaluate the performance of the combined network. 
We use implementation and pre-trained models provided by the authors. 
As in~\autoref{sec:exp-dcp}, we use the first 20 categories of ModelNet40 for training and validation and the last 20 categories solely for testing.
The training, validation, and testing splits are composed of 5,112, 1,202, and 1,266 models, respectively. 
As in~\cite{yew20CVPR}, we also report mean rotation error as in Eq. \eqref{eq:rot-error-iso} and mean translation error as in Eq. \eqref{eq:trans-error-p} with $p = 2$, marked as isotropic errors, and Chamfer distance \eqref{eq:chamfer-dist}. We report the experiment of point cloud alignment under Gaussian noise in the supplementary material and advance directly to the more challenging scenario of partial point cloud alignment.

\noindent{\bf Alignment with partial point clouds.} In this experiment, we evaluate partial point cloud registration. We randomly sample random half-space of the point cloud and retain only 70\% of the original point cloud, ensuring that there is a partial overlap in extent between point clouds. Simultaneously, we independently subsample 717 points of each half-space and add Gaussian noise $\mathcal{N}(0, 0.01^2)$ clamped at $[-0.05, 0.05]$.
RPM-Net uses two loss terms: $\mathcal{L}_\text{total} = \mathcal{L}_\text{reg} + \lambda \mathcal{L}_\text{inliers}$. While the $\mathcal{L}_\text{reg}$ term penalizes the pose error directly, the second $\lambda \mathcal{L}_\text{inliers}$ term acts directly on the correspondences. This term incentivizes the network to rank more matches as inliers. Since our layer also targets correspondence quality, we perform a test where we discard the inlier loss term.
%
\input{tables/partial_rpm}
As can be seen in \autoref{table:rpm-crop-performance}, we improve 
the rotation error by 0.137\degree and the translation error by 1e-3.
Contrary to the baseline, when our layer is included, it benefits from not having the inlier term acting directly on the correspondences. Without it, we manage to further lower rotation and translation errors.


\begin{table*}
    \begin{center}
    \resizebox{\linewidth}{!}{
    \begin{tabular}{lcccccccccr}
    \toprule
    Model \textbackslash (mean / std) & MSE($\mat R$)\degree & RMSE($\mat R$)\degree & MAE($\mat R$)\degree  & MSE($\vec t$) & RMSE($\vec t$) & MAE($\vec t$)\\
    \midrule
    DCP-v2 & 20.969088 / 1.439e+01 & 4.264873 / 1.667e+00 & 2.670795 / 9.765e-01 & \textbf{0.000047} / \textbf{5.962e-05} & \textbf{0.006250} / \textbf{2.900e-03} & \textbf{0.004555} / \textbf{2.153e-03}\\
    DCP-v2 + ours 5 all & \textbf{12.647069} / \textbf{1.046e+01} & \textbf{3.311812} / \textbf{1.296e+00} & \textbf{2.130985} / \textbf{7.879e-01} & 0.000052 / 6.235e-05 & 0.006532 / 3.039e-03 & 0.004757 / 2.291e-03 \\
    \bottomrule
    \end{tabular}
    }
    \end{center}
    \caption{
    Initialization test with Deep Closest Point on ModelNet40. Test with unseen point clouds where all the instances of the last 20 categories are held-out during training and only used for evaluation. The network is trained for 250 epochs.
    \label{table:exp-dcp-initialization}
    }
    \vspace{-1.5em}
    \end{table*}
    
    
    \begin{table}[t]
    \begin{center}
    \resizebox{\linewidth}{!}{%
        \begin{tabular}{lcccccccr}
        \toprule
        Model & RMSE($\mat R$)\degree & MAE($\mat R$)\degree  & RMSE($\vec t$) & MAE($\vec t$)\\ 
        \midrule
        DCP-v2 & 11.211800 & 8.554642 & 0.137432 & 0.105473 \\
        \textbf{DCP-v2 + ours}  & \textbf{10.232300} & \textbf{7.777227} & \textbf{0.128846} & \textbf{0.099061} \\
        \bottomrule
        \end{tabular}
    }
    \end{center}
    \caption{Deep Closest Point on 3DMatch with a 30\% minimum of surface overlap.\label{table:dcp:3dmatch}}
    \end{table}
    

\subsection{Deep Closest Point with 3DMatch Data}
\label{sec:exp-dcp-3dmatch}

The 3DMatch dataset is composed of 7317, 643 and 1861 point cloud pairs for training, validation and testing, respectively. Each fragment is downsampled to a point cloud of 1024 points through voxel grid average filtering \ie, all points inside a voxel are averaged to produce the resulting point for that voxel. Each point cloud pair is rescaled as to ensure that all points lie inside a unit ball norm, and each pair is guaranteed to have a minimum surface overlap of 30\%. We present our results in \autoref{table:dcp:3dmatch} and these show an improvement of 2.2\% and 1.7\% when normalized by the by the maximum magnitude of the rotation and translations
sampled, showing that out discoveries generalize to real data.

\subsection{Meaningful Improvements Beyond a Convenient Initialization}
\label{sec:initialization}

The inclusion of our layer produces improvements that occasionally can be marginal. To ensure that these are not attributed to a particular initialization we conduct an experiment where we train DCP 26 times with and without our layer, evaluating the mean and standard deviation of pose error over all runs. We conduct the experiment following the unseen categories protocol just as in previous sections. The results are presented in \autoref{table:exp-dcp-initialization}. 
Our method decreases the rotation error by 22\% while incurring in a residual increase in translation error.


\subsection{Measuring Correspondence Improvement}
\label{sec:exp-improved-correspondences}

\begin{figure}
\begin{center}
\includegraphics[width=0.8\linewidth]{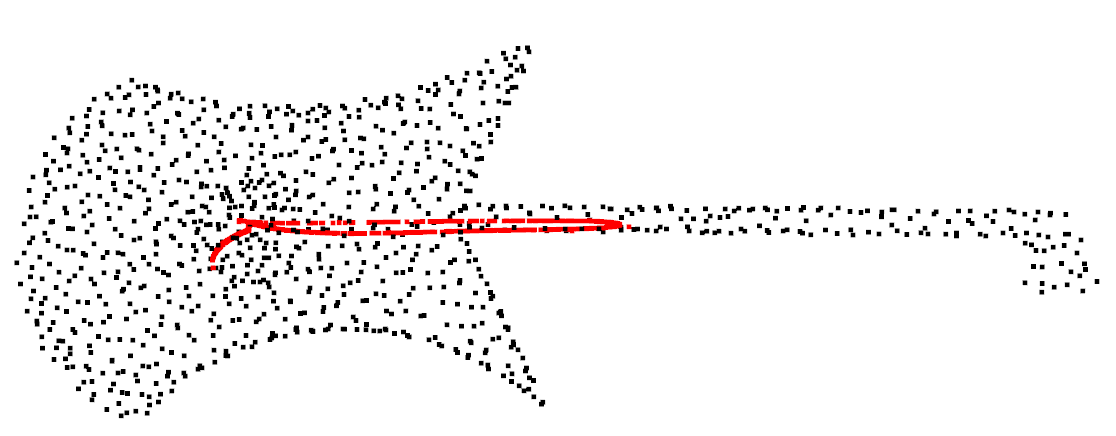}
\end{center}
  \caption{A qualitative example of the how correspondences are generated by Deep Closest Point. In the image we see point clouds of two different colors: black and red. We cherry-pick an example with the lowest pose estimation error, $\angle \Delta \mat R_\text{iso} = 0.2712\degree$, $\Delta \vec{t} = 0.0002$. \textbf{Black}: Point cloud generated by applying the ground-truth transformation to the source point cloud i.e., each point is given by $\vec{p}_{t_i} = \mat R_\text{gt} \vec{p}_{s_i} + \vec{t}_\text{gt}$. \textbf{Red}: The correspondences produced by the network to perform the registration task, where each point represents $\vec{p}'_{t_i}$.}
\label{fig:correspondences}
\vspace{-1.5em}
\end{figure}

In this section, we provide some intuition on why evaluating correspondence quality simply based on the Euclidean distance is misleading and only the error in the resulting rotation from Kabsch, accurately captures correspondence quality improvement. Due to space limitations, we further expand some of the results and ideas in the supplementary material.

\PAR{Correspondence Error is Misleading.} 
For each point $\vec{p}_{s_i}$ in the source point cloud, both DCP and RPM-Net regress the coordinates of its corresponding point, expressing it as $\vec{p}'_{t_i} = \sum_{j = 1}^N \alpha_{ij} \vec{p}_{t_j}$, where $\alpha_{ij}$ is the probability of point $\vec{p}_{s_i}$ matching $\vec{p}_{t_i}$. The mean subtracted version of these pairs of correspondences $\tilde{\vec{p}}_{s_i}$ and $\tilde{\vec{p}}'_{t_i}$ are used to compute $\mat H$ in Eq.~\eqref{eq:kabsch-input}.
To produce a correct rotation estimate, it is not necessary that $\|\tilde{\vec{p}}'_{t_i} - \mat R \tilde{\vec{p}}_{s_i}\|_{\forall i} = 0$. Kabsch is a method that is invariant to scale. Multiplying the source and target point clouds by arbitrary non-negative scalars will produce the same rotation matrix.
In an effort to evaluate the quality of correspondences based on the average point distance, we revisit the Gaussian noise experiment from \autoref{sec:exp-dcp}, where noise is added independently to one of the point clouds at test time. The results show that: the correspondence error is fairly high despite the low pose error; for a marginal improvement of 1\% in relative correspondence error, we obtain a 7\% improvement in rotation error, almost one order of magnitude above. This cements the idea that correspondence error does fully capture the quality of the correspondences established when these are used with Kabsch.

\PAR{In Defense of Pose Error as Evaluation Metric.} 
\autoref{fig:correspondences} shows a cherry picked example where the pose error is the smallest over the entire test set. Contrary to intuition, the regressed target points (red) hardly resemble the ground truth target points (black) and yet the network is still able to estimate an almost perfect pose. This confirms that a high positional error between regressed and ground truth target points does not imply a bad pose estimate.
However, both red and black point clouds have roughly similar principal directions and centroids and we conjecture that this is a sufficient condition for Kabsch to produce accurate estimates. Measuring a difference in orientation in principal directions and a difference in position between centroids is similar to computing a pose error. Therefore, we argue that pose error is a better metric to measure correspondence quality than correspondence positional error.

%% file: tables/partial_rpm.tex
%
%
\begin{table}
\small
\begin{center}
\setlength\tabcolsep{4.5pt}
\begin{tabularx}{\linewidth}{X | c c c c c}
  \toprule
  Method & \multicolumn{2}{c}{Anisotropic err.} & \multicolumn{2}{c}{Isotropic err.} & $\tilde{CD}$\\
   & (Rot.)\degree & (Trans.) & (Rot.)\degree & (Trans.) &  \\
  \midrule
  ICP         &  13.719  &  0.132  &  27.250  &  0.280  &  0.0153  \\
  RPM         &   9.771  &  0.092  &  19.551  &  0.212  &  0.0081  \\
  FGR         &  19.266  &  0.090  &  30.839  &  0.192  &  0.0119  \\
  PointNetLK  &  15.931  &  0.142  &  29.725  &  0.297  &  0.0235  \\
  DCP-v2      &  6.380   &  0.083  &  12.607  &  0.169   &  0.0113  \\
  \midrule
  RPM-Net        & 0.893      & 0.0087       & 1.712      & 0.018       &  \textbf{0.00085} \\
  RPM-Net + Ours & \textbf{0.826}      & 0.0081       & 1.575      & 0.017   &   \textbf{0.00085} \\
  \midrule
  RPM-Net$\dag$  & 0.993    & 0.0087    & 1.861    & 0.018   & 0.00099 \\
  RPM-Net + Ours$\dag$ & 0.872    & \textbf{0.0074}    & \textbf{1.554} & \textbf{0.015}   & 0.00088 \\
\bottomrule
\end{tabularx}
\end{center}
\caption{RPM-Net on ModelNet40: Performance on partially visible data with noise. The Chamfer distance using groundtruth transformations is 0.00055. \dag Models trained without the inlier term in the loss function.}
\label{table:rpm-crop-performance}
\end{table}

%% file: 05_supplementary_arxiv.tex
\appendix

\section{Introduction}

We address and expand certain aspects of the paper that were not relevant for its general understanding, but that give additional insight into our work and validate certain claims we made. We  provide proofs and detailed derivations whenever suitable, for some less obvious steps presented in the paper. 
The documents is organized into 7 main sections:
i) In \autoref{sec:derivation} we provide an in-depth derivation of our local approximate estimator;
ii) In \autoref{sec:linear-independence} we evaluate the redundancy of our constraints and show that once linearized around a rotation, the determinant constraint can be expressed as a linear combination of the orthogonality constraints. Furthermore, we also demonstrate the Linear Independence Constraint Qualification (LICQ) of the orthogonality constraints;
iii) In \autoref{sec:singularities} we demonstrate how the input matrix supplied to the \emph{rotation assembler} will never land on one of its singularities;
iv) In \autoref{sec:divergence} we provide insights into what makes the linear rotation estimator different from Kabsch, identify how these differences manifest themselves, as well as discuss how the problem geometry influences their occurrence;
v) In \autoref{sec:experiments} we include some additional experiments and ablations with Deep Closest Point and RPM-Net;
vi) In \autoref{sec:correspondences} we expand on our claim that the most accurate way of establishing correspondence quality in Kabsch is by looking at pose error.

\section{Detailed Derivation of the Linearized Local Approximation Estimator}
\label{sec:derivation}

Given a set of correspondences between the source and target point clouds $\mat P_s, \mat P_t \in \R^{N\times 3}$ our aim is to find a rigid transformation ($\mat R$, $\vec t$) that minimizes the following error:
\begin{subequations}
\label{eq:original_s}
\begin{eqnarray}
\argmin_{\mat R, \vec t} & \sum_{i=1}^N w_i \|\vec{p_t}_i  - \mat R \vec{p_s}_i - \vec t\|^2 \label{eq:original-cost_s}\\
\st & \mat R \in SO(3),
\end{eqnarray}
\end{subequations}
where $\vec w \in \R_+^N$ represents a (n optionally supplied) set of weights and $\vec{p_s}_i, \vec{p_t}_i \in \R^3$ are individual points of the source and target point clouds.
Given an optimal rotation matrix $\mat R$, the problem reduces itself to a linear regression, where we can simply extract the optimal translation vector $\vec t$ in a closed-form as:
\begin{equation}
    \vec t^* = \frac{\sum_{i=1}^N w_i(\vec{p_t}_i  - \mat R \vec{p_s}_i)}{\sum_{i=1}^N w_i} = \bar{\vec{p_t}} - \mat R \bar{\vec{p_s}}, \label{eq:optimal-t_s}
\end{equation}
where $\bar{\vec{p_t}}$ and $\bar{\vec{p_s}}$ represent the weighted means of the points for each point cloud. 
We further define $\tilde{\vec{p_t}}_i$ and $\tilde{\vec{p_s}}_i$ as the mean-subtracted versions of $\vec{p_t}_i$ and $\vec{p_s}_i$, such that $\tilde{\vec{p_s}}_i = \vec{p_s}_i - \bar{\vec{p_s}}$ and $\tilde{\vec{p_t}}_i = \vec{p_t}_i - \bar{\vec{p_t}}$. We can then factor out the translation component and Eq.~\eqref{eq:original_s} can be formulated entirely with the respect to the rotation. Back-substituting Eq.~\eqref{eq:optimal-t_s} yields the following simplification:
\begin{subequations}
\label{eq:simplified_s}
\begin{eqnarray}
\argmin_{\mat R} & \sum_{i=1}^N w_i \|\tilde{\vec{p_t}}_i  - \mat R \tilde{\vec{p_s}}_i\|^2 \label{eq:simplified-cost_s}\\
\st & \mat R \in SO(3) \label{eq:simplified-constraints_s}.
\end{eqnarray}
\end{subequations}
The membership of $\mat R$ in $SO(3)$ can be expressed as: 
\begin{subequations}
\label{eq:constraints_s}
\begin{eqnarray}
\mat R^\top \mat R &=& \mat I_3 \label{eq:orthogonality_s}\\
\det \mat R &=& 1, \label{eq:const-det_s}
\end{eqnarray}
\end{subequations}
where $\mat I_3$ is a $3\times3$ identity matrix.
These are quadratic and cubic equality constraints, respectively. Eq.~\eqref{eq:orthogonality_s} supplies six constraints and Eq.~\eqref{eq:const-det_s} an additional one. However, when evaluating the first-order terms of the its Taylor expansion around a rotation matrix, only the orthogonality constraints respect the LICQ. We can optionally drop either one of the orthogonality constraints or the determinant constraint. We drop the latter to simplify the formulation. We expand this further in \autoref{sec:linear-independence}.

\PAR{Linearization of Constraints.}
Denoting our prior rotation estimate as $\mat R_{t-1}$,
we linearize Eq.~\eqref{eq:orthogonality_s} around it, only taking into consideration the upper triangle section of the constraints' matrix.
We define matrix $c(\mat R) = \mat R^\top \mat R - \mat I$ such that $c(\mat R): \R^{3\times3} \to \R^{3\times3}$ and refer to $c_{ij}(\mat R)$ as the element in the $i$-th row and $j$-th column, defined as $c_{ij}(\mat R) = \vec e_i^\top (\mat R^\top \mat R - \mat I) \vec e_j$. 
The variables $\vec e_i, \vec e_j \in \R^3$ are Euclidean bases, vectors of zeros with a single element equal to one at the $i$-th and $j$-th elements, respectively. To linearize it, let us formulate a first-order Taylor expansion for each element in $c(\mat R)$, around an initial estimate $\mat R_{t-1}$:
\begin{equation}
    c_{ij}^{(1)}(\mat R, \mat R_{t-1}) = c_{ij}(\mat R_{t-1}) + \tr\left(\frac{\partial c_{ij}(\mat R_{t-1})}{\partial \mat R}^\top (\mat R - \mat R_{t-1})\right),
\end{equation}
where the superscript ${}^{(1)}$ represents the first-order Taylor approximation. The derivative $\frac{\partial c_{ij}(\mat R)}{\partial \mat R}$ is computed as follows:
\begin{subequations}
\begin{align}
    \frac{\partial c_{ij}(\mat R)}{\partial \mat R} &= \frac{\partial }{\partial \mat R}  \vec e_i^\top (\mat R^\top \mat R - \mat I) \vec e_j \\
    &= \frac{\partial }{\partial \mat R}  \vec e_i^\top \mat R^\top \mat R\vec e_j  \\
    &= \mat R\vec e_j \vec e_i^\top  +  \mat R\vec e_i \vec e_j^\top  .
\end{align}
\label{eq:constraint-derivative}
\end{subequations}
Substituting it back yields,
\begin{eqnarray}
c_{ij}^{(1)}(\mat R, \mat R_{t-1}) &= c_{ij}(\mat R_{t-1}) + \tr\left(\mat E^\mathbb{S}_{ij} \mat R_{t-1}^\top (\mat R -  \mat R_{t-1})\right) \label{eq:constraint-orthogonal-linear_s}\\
\text{for} & i = 1, \dots, 3; \, j=i, \dots, 3, \nonumber
\end{eqnarray}
where $\mat E^\mathbb{S}_{ij} = \vec e_i \vec e^\top_j + \vec e_j \vec e^\top_i = \mat E_{ij} + \mat E_{ji} \in \mathbb{S}^3$ and $\mat E_{ij} = \vec e_i \vec e_j^\top$.

\PAR{Langrangian Formulation.}
After the relaxation and linearization of our constraints, we now have an optimization problem with a quadratic cost function and linear constraints. Then, we can enforce the (linearized) equality  constraints in Eq.~\eqref{eq:constraint-orthogonal-linear_s} using the method of Lagrange multipliers. Thus, we obtain a closed-form solution that can be formulated as a linear system of equations. We write the Lagrangian of Eq.~\eqref{eq:simplified-cost_s} as: 
\begin{equation}
\mathcal{L}(\mat R, \mat R_{t-1}) = \sum_{i=1}^N \frac{w_i}{2} \|\tilde{\vec{p_t}}_i  - \mat R \tilde{\vec{p_s}}_i\|^2 + \sum_{k=1}^6 \lambda_k c^{(1)}_k(\mat R, \mat R_{t-1}), \label{eq:lagrange_s}
\end{equation}
where indices $ij$ previously used to specify the row and column of the constraints $c_{ij}^{(1)}(\mat R, \mat R_{t-1})$, are now replaced by the single index $k$, iterating over the upper triangular part of the matrix:
\begin{align}
    c^{(1)}(\mat R, \mat R_{t-1}) &= \underbrace{\begin{bmatrix}
    c_{11} & c_{12} & c_{13} \\
        & c_{22} & c_{23} \\
        &   & c_{33}
    \end{bmatrix}}_{ij}
    = \underbrace{\begin{bmatrix}
    c_{1} & c_{2} & c_{4} \\
        & c_{3} & c_{5} \\
        &   & c_{6}
    \end{bmatrix}}_{k}.
\end{align}
The variables $\lambda_k$ represent the Lagrange multipliers of the constraints. The minimum to the original constrained problem is guaranteed to be a stationary point of the Lagrangian. Thus, we compute the gradient of the Lagrangian and set it to 0 to look for possible optimal candidates. We start by fully expanding all terms in Eq.~\eqref{eq:lagrange_s}:
\begin{align}
\mathcal{L}(\mat R, \mat R_{t-1}) &= \sum_{i=1}^N \frac{w_i}{2} \tr(\tilde{\vec{p_t}}_i^\top \tilde{\vec{p_t}}_i  - 2 \tilde{\vec{p_t}}_i^\top \mat R \tilde{\vec{p_s}}_i + \tilde{\vec{p_s}}_i^\top \mat R^\top \mat R \tilde{\vec{p_s}}_i) \nonumber \\
&+ \sum_{k=1}^6 \lambda_k \left(c_k(\mat R_{t-1}) + \tr\left(\mat E^\mathbb{S}_k \mat R_{t-1}^\top (\mat R -  \mat R_{t-1})\right)\right), \label{eq:lagrange-explanded}
\end{align}
followed by computing its partial derivatives:
\begin{subequations}
\begin{align}
\frac{\partial \mathcal{L}(\mat R, \mat R_{t-1})}{\partial \mat R} &= \sum_{i=1}^N w_i (\mat R \tilde{\vec{p_s}}_i \tilde{\vec{p_s}}_i^\top - \tilde{\vec{p_t}}_i \tilde{\vec{p_s}}_i^\top) \nonumber \\
&+ \sum_{k=1}^6 \lambda_k  \mat R_{t-1} \mat E^\mathbb{S}_k \label{eq:gradient-r}\\
\frac{\partial \mathcal{L}(\mat R, \mat R_{t-1})}{\partial \lambda_k} &= c_k(\mat R_{t-1}) + \tr\left(\mat E^\mathbb{S}_k \mat R_{t-1}^\top (\mat R -  \mat R_{t-1})\right). \label{eq:gradient-lambda}
\end{align}
\label{eq:gradient}
\end{subequations}
We find the stationary points by finding the values of $\mat R$ and $\lambda_k$ that verify $\nabla \mathcal{L} = 0$. These equations are linear \wrt to the variables $\mat R$ and $\lambda_k$, meaning we can estimate them by solving a linear system of equations. The system needs to be rewritten so that a linear solver can estimate estimate the unknowns $\begin{bmatrix}
\veco(\mat R)^\top & \vec \lambda^\top
\end{bmatrix}^\top$. The operator $\veco$, represents a column-wise vectorization operator and $\vec \lambda = [\lambda_1, \dots, \lambda_6]^\top$. Let us also define 
$\vec r = \veco(\mat R)$ for brevity. The optimal $\mat R$ and $\vec \lambda$ are determined by solving a linear system of the form: 
\begin{equation}
\begin{bmatrix}
\mat A & \mat B \\ \mat B^\top & 0 \end{bmatrix} \begin{bmatrix}
\vec r \\ \vec \lambda
\end{bmatrix} = \begin{bmatrix}
\vec d_{\mat R} \\
\vec d_{\vec \lambda}
\end{bmatrix}. \label{eq:linear-system}
\end{equation}
We derive $\mat A$, $\mat B$, $\vec d_{\mat R}$ and $\vec d_{\vec \lambda}$ in the following paragraphs.

\PAR{Partial Derivative \wrt $\mat R$.}
We need to rearrange the terms in Eq.~\eqref{eq:gradient} in order to now express them \wrt $\vec r$. Starting with Eq.~\eqref{eq:gradient-r}, $\frac{\partial \mathcal{L}(\mat R, \mat R_{t-1})}{\partial \mat R} : \R^{3\times3} \to \R^{3\times3}$,  we need to access individual elements of this matrix in order to vectorize $\mat R$. Applying a similar step to Eq.~\eqref{eq:constraint-derivative}, we formulate
\begin{equation}
    \frac{\partial \mathcal{L}(\mat R, \mat R_{t-1})}{\partial \mat R}_{mn} = \vec e_m^\top \frac{\partial \mathcal{L}(\mat R, \mat R_{t-1})}{\partial \mat R} \vec e_n,
\end{equation}
for $m = {1, \dots, 3}$ and $n = {1, \dots, 3}$.
We introduce the following identity $\tr(\mat A^\top \mat B) = \veco(\mat A)^\top \veco(\mat B)$.
Focusing only in the terms that depend on $\mat R$, we have:
\begin{align}
\frac{\partial \mathcal{L}_{\mat R}(\mat R, \mat R_{t-1})}{\partial \mat R} &= \vec e_m^\top \mat R \sum_{i=1}^N w_i  \tilde{\vec{p_s}}_i \tilde{\vec{p_s}}_i^\top \vec e_n \\
&= \tr \left( \vec e_m^\top \mat R \sum_{i=1}^N w_i  \tilde{\vec{p_s}}_i \tilde{\vec{p_s}}_i^\top \vec e_n \right) \\
&= \tr \left( \sum_{i=1}^N w_i  \tilde{\vec{p_s}}_i \tilde{\vec{p_s}}_i^\top \vec e_n \vec e_m^\top \mat R  \right) \\
&= \tr \left( \left(\vec e_m \vec e_n^\top \sum_{i=1}^N w_i  \tilde{\vec{p_s}}_i \tilde{\vec{p_s}}_i^\top\right)^\top \mat R  \right) \\
&= \tr \left( \left(\mat E_{mn} \sum_{i=1}^N w_i  \tilde{\vec{p_s}}_i \tilde{\vec{p_s}}_i^\top\right)^\top \mat R  \right) \\
&= \vec a_{mn}^\top \vec r,
\end{align}
with $\vec a_{mn} = \veco\left(\mat E_{mn} \sum_{i=1}^N w_i  \tilde{\vec{p_s}}_i \tilde{\vec{p_s}}_i^\top\right)$. This generates an equation for each element of $\frac{\partial \mathcal{L}(\mat R, \mat R_{t-1})}{\partial \mat R}$. However, we still need to stack these equations as rows of a matrix, in a suitable form for a linear solver. We do so by vectorizing $\frac{\partial \mathcal{L}(\mat R, \mat R_{t-1})}{\partial \mat R}$ following a column-wise order. This composes matrix $\mat A$ as presented in the paper:
\begin{equation}
    \mat A = \begin{bmatrix}
    \vec a_{11} & \vec a_{12} & \vec a_{13} & \vec a_{21} & \dots & \vec a_{33}
    \end{bmatrix}^\top. \label{eq:linear-system-a}
\end{equation}
In the main paper, we express the different rows of $\mat A$ according to a single index $r$. The mapping between indices $mn$ and $r$ is given by,
\begin{equation}
\underbrace{
    \begin{bmatrix}
    a_{11} & a_{12} & a_{13} \\
    a_{21} & a_{22} & a_{23} \\
    a_{31} & a_{32} & a_{33}
    \end{bmatrix}}_{mn} = \underbrace{
    \begin{bmatrix}
    a_{1} & a_{4} & a_{7} \\
    a_{2} & a_{5} & a_{8} \\
    a_{3} & a_{6} & a_{9}
    \end{bmatrix}}_{r}.
\end{equation}

Next we look at $\frac{\partial \mathcal{L}_{\lambda_k}(\mat R, \mat R_{t-1})}{\partial \mat R}$ \ie, all terms from the partial derivative that depend on $\lambda_k$:
\begin{equation}
    \frac{\partial \mathcal{L}_{\lambda_k}(\mat R, \mat R_{t-1})}{\partial \mat R} = \lambda_k  \mat R_{t-1} \mat E^\mathbb{S}_k.
\end{equation}
The terms $\lambda_k$ are already scalar so we only need to vectorize the expression in order stack all equation like we did for Eq.~\eqref{eq:linear-system-a}, resulting in the following definitions
\begin{align}
    \vec b_k &= \veco(\mat R_{t-1} \mat E^\mathbb{S}_k) \label{eq:linear-system-b}\\
    \mat B &= \begin{bmatrix}
    \vec b_1 & \dots & \vec b_6
    \end{bmatrix}.
\end{align}

All that remains are the constant terms
\begin{equation}
    \frac{\partial \mathcal{L}_{\text{constant}}(\mat R, \mat R_{t-1})}{\partial \mat R} = \sum_{i=1}^N w_i  \tilde{\vec{p_t}}_i \tilde{\vec{p_s}}_i^\top,
\end{equation}
that similar to Eqs.~\eqref{eq:linear-system-a} \eqref{eq:linear-system-b}, are also vectorized:
\begin{equation}
    \vec d_{\mat R} = \veco\left(\sum_{i=1}^N w_i  \tilde{\vec{p_t}}_i \tilde{\vec{p_s}}_i^\top\right).
\end{equation}
This concludes the upper part of the linear system of equations.

\PAR{Partial Derivative \wrt $\lambda_k$.}

Every partial derivate \wrt $\lambda_k$, with $k = 1, \dots, 6$, contributes with an equation to the linear system. Addressing first, the terms that depend on $\mat R$, we have 
\begin{align}
    \frac{\partial \mathcal{L}_{\mat R}(\mat R, \mat R_{t-1})}{\partial \lambda_k} &= \tr\left(\mat E^\mathbb{S}_k \mat R_{t-1}^\top \mat R\right) \\
    &= \tr\left((\mat R_{t-1} \mat E^\mathbb{S}_k)^\top \mat R\right) \\
    &= \veco\left(\mat R_{t-1} \mat E^\mathbb{S}_k\right)^\top \vec r \\
    &= \vec b_k^\top \vec r,
\end{align}
now expressed in terms of $\vec r$. Performing the same operation for every value of $k$ and stacking $\vec b_k^\top$ as rows will produce the matrix $\mat B^\top$.

The only terms remaining are
\begin{align}
    \frac{\partial \mathcal{L}_{\text{constant}}(\mat R, \mat R_{t-1})}{\partial \lambda_k} &= \tr\left(\mat E^\mathbb{S}_k \mat R_{t-1}^\top \mat R_{t-1}\right) - c_k(\mat R_{t-1}) \\
    &= \tr\left(\mat E^\mathbb{S}_k\right) - c_k(\mat R_{t-1}) \\
    &= d_{\lambda_k}.
\end{align}
Stacking $d_{\lambda_k}$ for every possible value of $k$ produces the vector $\vec d_{\vec \lambda} = [d_{\lambda_1}, \dots, d_{\lambda_6}]^\top$.

\section{Linear Independence Constraint Qualification}
\label{sec:linear-independence}

Our optimization problem is governed by rotation constraints as expressed in Eqs.~\eqref{eq:constraints_s}. 
In order to be able to employ the method of Lagrange Multipliers, our problem needs to verify the LICQ. In our particular case, ensuring the LICQ is what guarantees that the linear system of equations in Eq.~\eqref{eq:linear-system} can be inverted and we can retrieve a unique solution.
Contrary to what was done in the previous section, to study the LICQ, it is convenient to express constraints \wrt to the column vectors of $\mat R$. To do so, let us define vectors $\vec r_1$, $\vec r_2$ and $\vec r_3$, such that
\begin{equation}
    \mat R = \begin{bmatrix}
    \vrule & \vrule & \vrule \\
    \vec r_1 & \vec r_2 & \vec r_3 \\
    \vrule & \vrule & \vrule
    \end{bmatrix},
\end{equation}
where $\vec r_i \in \R^3$, for $i = \{1, 2, 3\}$. With these new variables, we now represent Eqs.~\eqref{eq:constraints_s} as:
\begin{subequations}
\begin{align}
    &\vec r_i^\top \vec r_i = 1 \quad \text{for } i = \{1, 2, 3\}\\
    &\vec r_i^\top \vec r_j = 0 \quad \text{for } (i,j) = \{(1, 2), (2, 3), (3, 1)\} \\
    &\vec r_1^\top \lfloor \vec r_2 \rfloor_{\times} \vec r_3 = 1 \quad \text{(determinant)}. \label{eq:constraint-col-determinant}
\end{align}
\end{subequations}
The operator $\lfloor \cdot \rfloor_{\times}$ takes a vector in $\R^3$ and produces the skew symmetric matrix
\begin{equation}
    \lfloor \vec v \rfloor_\times = \begin{bmatrix}
    0 & -v_z  & v_y \\
    v_z & 0 & -v_x \\
    -v_y & v_x & 0
    \end{bmatrix}.
\end{equation}
This matrix can be used to represent a vector cross-product in matrix form, e.g. $\vec v_1 \times \vec v_2 = \lfloor \vec v_1 \rfloor_\times \vec v_2$. The determinant constraint in Eq.~\eqref{eq:constraint-col-determinant} is expressed as the scalar triple product of the column vectors. A scalar triple product is a product of the form $\vec v_1 \cdot (\vec v_2 \times \vec v_3)$. The product does not change with a circular shift of the vectors, meaning the determinant of $\mat R$ can also be represented as $\vec r_2^\top \lfloor \vec r_3 \rfloor_{\times} \vec r_1$ or $\vec r_3^\top \lfloor \vec r_1 \rfloor_{\times} \vec r_2$.

We write the first-order Taylor expansion for each constraint, now formulated \wrt column vectors, assuming the form:
\begin{equation}
    c^{(1)}(\vec r_i, \vec r_{i_{t - 1}}) = c(\vec r_{i_{t - 1}}) + \frac{\partial c(\vec r_{i_{t - 1}})}{\partial \vec r_i}^\top (\vec r_i - \vec r_{i_{t-1}}).
\end{equation}
The method of Lagrange multipliers requires that equality constraints are expressed as functions that are equal to 0. At the same time, because the linearization point $\mat R_{t-1}$ will always be a rotation matrix, its columns will always ensure that $c(\vec r_{i_{t-1}}) = 0$.

\begin{figure}
\begin{center}
\includegraphics[width=0.98\linewidth]{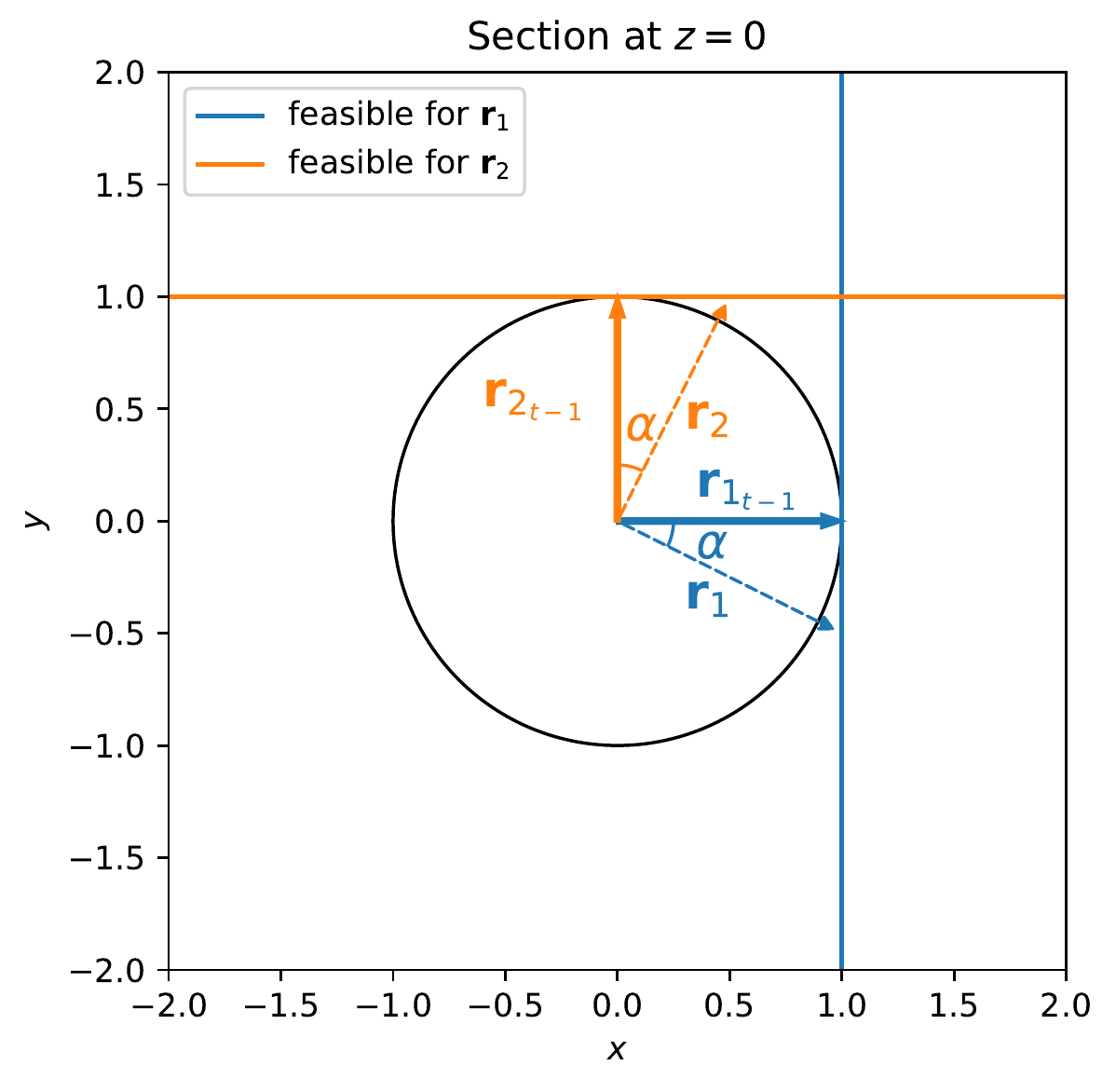}
\end{center}
  \caption{This figure depicts how rotation (matrix) orthogonality constraints manifest themselves in space once they are linearized. The figure shows a cross section at $z = 0$ and only the first two column vectors of the rotation matrices $\mat R$ and $\mat R_{t -1}$ are represented. The frame of reference is aligned with the bases of $\mat R_{t -1}$. The (blue) vertical and (orange) horizontal lines represent the 3D planes created from linearizing the vector's unit norm constraints. They show that the resultant column vectors of the matrix our linear estimator produces, will always have a norm larger or equal to one. Simultaneously, we can also see that the \emph{projections} of $\vec r_1$ and $\vec r_2$ to the plane formed by vectors $\vec r_{1_{t-1}}$ and $\vec r_{2_{t-1}}$, in this image represented as $z = 0$, will also be orthogonal to each other. However, since the vectors $\vec r_1$ and $\vec r_2$ are not restricted to $z = 0$, there is no guarantee that they are orthogonal.}
\label{fig:feasible-set}
\vspace{-1.5em}
\end{figure}

\PAR{Unit Norm.}
We look first into the (matrix) orthogonality constraints that place requirements in the norm of the column vectors.  Following this requirement, we define
\begin{equation}
    c_{n_i} = \vec r_i^\top \vec r_i - 1 \quad \text{for } i = \{1, 2, 3\}.
\end{equation}
The corresponding derivative at the linearization point is given by
\begin{equation}
    \frac{\partial c_{n_i}(\vec r_{i_{t-1}})}{\partial \vec r_i} = 2 \vec r_{i_{t-1}},
\end{equation}
resulting in the following linearized constraint
\begin{align}
    c_{n_i}^{(1)} &= 2\vec r_{i_{t-1}}^\top (\vec r_{i} - \vec r_{i_{t-1}})\\
    &= 2 (\vec r_{i_{t-1}}^\top \vec r_i - 1) \quad \text{for } i = \{1, 2, 3\}.
\end{align}
For brevity, we shall omit the explicit dependency of $c_{n_i}$ \wrt $\vec r_i$ and $\vec r_{i_{t-1}}$, as well as in all constraints mentioned henceforth. This constraint forces the vector $\vec r_i$ to belong to a tangent plane to the unit sphere, with the plane defined by the normal $\vec r_{i_{t-1}}$ as displayed in \autoref{fig:feasible-set}.

\PAR{Orthogonality.}
The column (and row) vectors of a rotation matrix are all perpendicular to each other. This constraint is expressed as
\begin{equation}
    c_{o_{ij}} = \vec r_i^\top \vec r_j \quad \text{for } (i,j) = \{(1, 2), (2, 3), (3, 1)\}.
\end{equation}
The corresponding derivatives at the linearization point are given by
\begin{equation}
    \frac{\partial c_{o_{ij}}}{\partial \vec r_i} = \vec r_{j_{t-1}},  \quad \frac{\partial c_{o_{ij}}}{\partial \vec r_j} = \vec r_{i_{t-1}},
\end{equation}
resulting in the following linearized constraint
\begin{equation}
    c_{o_{ij}}^{(1)} = \vec r_{j_{t-1}}^\top \vec r_i + \vec r_{i_{t-1}}^\top \vec r_j \quad \text{for } (i,j) = \{(1, 2), (2, 3), (3, 1)\}.
\end{equation}
Please refer to \autoref{fig:feasible-set} for additional insights on how these constraints look in space.

\PAR{Unit Determinant.}
The determinant constraint of a rotation matrix is given by
\begin{align}
    c_d &= \vec r_1^\top \lfloor \vec r_2 \rfloor_\times \vec r_3 - 1 \\
    &= \vec r_2^\top \lfloor \vec r_3 \rfloor_\times \vec r_1 - 1 \\
    &= \vec r_3^\top \lfloor \vec r_1 \rfloor_\times \vec r_2 - 1.
\end{align}
We will make use of all three variants to compute the different partial derivatives. The partial derivatives at the linearization point are
\begin{align}
    \frac{\partial c_d}{\partial \vec r_1} &= \lfloor \vec r_{2_{t-1}} \rfloor_\times \vec r_{3_{t-1}} \\
    &= \vec r_{1_{t-1}} \\
    \frac{\partial c_d}{\partial \vec r_2} &= \lfloor \vec r_{3_{t-1}} \rfloor_\times \vec r_{1_{t-1}} \\
    &= \vec r_{2_{t-1}} \\
    \frac{\partial c_d}{\partial \vec r_3} &= \lfloor \vec r_{1_{t-1}} \rfloor_\times \vec r_{2_{t-1}} \\
    &= \vec r_{3_{t-1}},
\end{align}
resulting in the following linearized constraint
\begin{align}
    c_d^{(1)} &= \sum_{i=1}^3\vec r_{i_{t-1}}^\top (\vec r_{i} - \vec r_{i_{t-1}}) \\
    &= \sum_{i=1}^3\vec r_{i_{t-1}}^\top \vec r_{i} - 3.
\end{align}
The determinant constraint can be written as
\begin{equation}
    c_d^{(1)} = \frac{1}{2} \sum_{i=1}^3 c_{n_i}^{(1)},
\end{equation}
implying this constraint is linearly dependent and as such it is removed from the final formulation.

\PAR{Linear Independence Constraint Qualification.}
In order for the method of Lagrange Multipliers to produce a single solution, our constraints need to verify the Linear Independence Constraint Qualification (LICQ). The LICQ requires all gradients of the constraints to be linearly independent at the optimum $\mat R^*$, assuming one exists. Since the determinant constraint could be formulated as a linear combinations of the orthogonality ones, it was dropped. If a constraint is linearly dependent, its gradients will also be. If we only consider the (matrix) orthogonality constraints and we stack all their gradients into a single matrix it yields
\begin{equation}
    \mat C =
    \begin{bmatrix}
    2 \vec r_{1_{t-1}} & 0 & 0 & \vec r_{2_{t-1}} & \vec r_{3_{t-1}} & 0\\
    0 & 2 \vec r_{2_{t-1}} & 0 & \vec r_{1_{t-1}} & 0 & \vec r_{3_{t-1}} \\
    0 & 0 & 2 \vec r_{3_{t-1}} & 0 & \vec r_{1_{t-1}} & \vec r_{2_{t-1}}
    \end{bmatrix},
\end{equation}
where $\mat C \in \R^{9\times6}$. Each column $\vec c_i \in \R^9$ of $\mat C$, represents a gradient of one of the constraints. These vectors are constant because the constraints are linear. Therefore, if we ensure they are linearly independent, this also holds at the optimum $\mat R^*$. To guarantee linear independence, the columns of $\mat C$ need to satisfy the following property:
\begin{equation}
    \sum_{i=1}^6 \alpha_i \vec c_i = 0 \to \forall \, 1 \leq i \leq 6: \alpha_i = 0.
\end{equation}
The entries in $\mat C$ consist of the column vectors of matrix $\mat R_{t-1} = \begin{bmatrix}
\vec r_{1_{t-1}} & \vec r_{2_{t-1}} & \vec r_{1_{t-1}}
\end{bmatrix} \in SO(3)$. This matrix is the solution at iteration $t - 1$ and hence satisfies the properties in Eqs. \eqref{eq:orthogonality_s} and \eqref{eq:const-det_s}. Eq. \eqref{eq:orthogonality_s} ensures $\mat R_{t-1}$ is orthogonal. This implies that its column (and row) vectors are linearly independent and hence we are given that for $\forall \beta_1, \beta_2, \beta_3 \in \R$ with $\beta_1 \vec r_{1_{t-1}} + \beta_2 \vec r_{2_{t-1}} + \beta_3 \vec r_{3_{t-1}} = 0 \Rightarrow \beta_1 = \beta_2 = \beta_3 = 0$. Now assume that we are given such $\alpha_1, \dots, \alpha_6 \in \R$ with $\alpha_1 \vec c_1 + \dots + \alpha_6 \vec c_6 = 0$. We obtain the following three equations:
\begin{align}
    2 \alpha_1 \vec r_{1_{t-1}} + \alpha_4 \vec r_{2_{t-1}} + \alpha_5 \vec r_{3_{t-1}} &= \begin{bmatrix}0 \\ 0 \\ 0\end{bmatrix} \label{eq:lin-ind-1}\\
    2 \alpha_2 \vec r_{2_{t-1}} + \alpha_4 \vec r_{1_{t-1}} + \alpha_6 \vec r_{3_{t-1}} &= \begin{bmatrix}0 \\ 0 \\ 0\end{bmatrix} \label{eq:lin-ind-2}\\
    2 \alpha_3 \vec r_{3_{t-1}} + \alpha_5 \vec r_{1_{t-1}} + \alpha_6 \vec r_{2_{t-1}} &= \begin{bmatrix}0 \\ 0 \\ 0\end{bmatrix}. \label{eq:lin-ind-3}
\end{align}
Since we know that the columns vectors $\vec r_{1_{t-1}}$ with $1 \leq i \leq 3$ are linear independent, we can deduce that
\begin{align}
    \eqref{eq:lin-ind-1} &\to \alpha_1 = \alpha_4 = \alpha_5 = 0 \\
    \eqref{eq:lin-ind-2} &\to \alpha_2 = \alpha_4 = \alpha_6 = 0 \\
    \eqref{eq:lin-ind-3} &\to \alpha_4 = \alpha_5 = \alpha_6 = 0,
\end{align}
and thus we obtain that $\forall \, 1 \leq i \leq 6 : \alpha_i = 0$.

\section{Orthogonalization Singularities}
\label{sec:singularities}

In the paper, we make the claim that during the orthogonalization steps of our \emph{rotation assembler} stage, the estimates produced by our linear estimator will not lie close to the singularities of this operation. We present once more the steps involved into turning the estimates back into rotation matrices.
Assume that $\mat R'$ is a $3\times3$ input matrix to this stage, resulting from the output of the linear estimator and formed by the columns $\vec r'_1, \vec r'_2, \vec r'_3 \in \R^3$. This matrix is constructed in the following way:
\begin{eqnarray}
    \vec r_1 &=& \frac{\vec r'_1}{\|\vec r'_1\|}, \label{eq:rot-composition-r1_s}\\
    \vec r_2 &=& \frac{(\mat I_3 - \vec r_1 \vec r_1^\top)\vec r'_2}{\|(\mat I_3 - \vec r_1 \vec r_1^\top)\vec r'_2\|}, \label{eq:rot-composition-r2_s}\\
    \vec r_3 &=& \vec r_1 \times \vec r_2. \label{eq:rot-composition-r3_s}
\end{eqnarray}

To avoid the singularities associated with the denominator terms of Eqs.~\eqref{eq:rot-composition-r1_s} and \eqref{eq:rot-composition-r2_s}, we need to ensure that both $\|\vec r'_1\| > 0$  and $\|(\mat I_3 - \vec r_1 \vec r_1^\top)\vec r'_2\| > 0$. This is accomplished by guaranteeing that the input column vectors have a positive norm, something that is directly observable in \autoref{fig:feasible-set}, and by guaranteeing that $\vec r'_1$ and $\vec r'_2$ are not collinear.

\PAR{Positive Vector Norm.}
To show that all column vectors have a positive norm, we recover the linearized norm constraints
\begin{equation}
    \vec r_{i_{t-1}}^\top \vec r'_i = 1 \quad \text{for } i = \{1, 2, 3\}. \label{eq:linearized-norm-constraint-2}
\end{equation}
From here the we perform the following manipulation
\begin{align}
    & \quad \vec r_{i_{t-1}}^\top \vec r'_i = 1 \\
    \Leftrightarrow & \quad -2 \vec r_{i_{t-1}}^\top \vec r'_i = -2 \\
    \Leftrightarrow & \quad \|\vec r_{i_{t-1}}\|^2 -2 \vec r_{i_{t-1}}^\top \vec r'_i + \|\vec r'_i\|^2 = \|\vec r_{i_{t-1}}\|^2 -2 + \|\vec r'_i\|^2 \\
    \Leftrightarrow & \quad \|\vec r'_i - \vec r_{i_{t-1}}\|^2 = \|\vec r'_i\|^2 - 1  \\
    \Leftrightarrow & \quad 1 \leq \|\vec r'_i\|,
\end{align}
confirming the intuition from \autoref{fig:feasible-set}.

\PAR{Non-collinear Vectors.}
We show that $\vec r'_1$ and $\vec r'_2$ cannot be collinear by contradiction. If $\vec r'_1$ were $\vec r'_2$ indeed collinear, one would be able to write $\vec r'_2 = a \vec r'_1$, with $a$ representing an arbitrary non-null scalar in $\R$. However, if we substitute this relation into the linearized (vector) orthogonality constraints, it yields
\begin{align}
    & \quad \vec r_{2_{t-1}}^\top \vec r'_1 + \vec r_{1_{t-1}}^\top \vec r'_2 = 0 \\
    \Leftrightarrow & \quad \frac{1}{a}\vec r_{2_{t-1}}^\top \vec r'_2 + a \vec r_{1_{t-1}}^\top \vec r'_1 = 0 \\
    \Leftrightarrow & \quad \frac{1}{a}\underbrace{\vec r_{2_{t-1}}^\top \vec r'_2}_{\overset{\eqref{eq:linearized-norm-constraint-2}}{=}1} + a \underbrace{\vec r_{1_{t-1}}^\top \vec r'_1}_{\overset{\eqref{eq:linearized-norm-constraint-2}}{=}1} = 0 \\
    \Leftrightarrow & \quad \frac{1}{a} + a = 0 \\
    \Leftrightarrow & \quad a^2 = -1.
\end{align}
There is no $a \in \R$ that satisfies the equation above, contradicting the original assumption that $\vec r'_1$ and $\vec r'_2$ are collinear.

\section{Understanding the Differences Between Estimators}
\label{sec:divergence}

As mentioned in the main paper, the estimator we propose solves a very similar problem to Kabsch, minimizing the same correspondence loss, but under a different set of constraints: a linear approximation of the original second-order equality constraints. A network trained with and without our additional layers will learn a different set of parameters and will have different registration performance, a byproduct of the different gradients that our extra layers produce. To monitor gradient differences, we revisit Deep Closest Point~\cite{wang19ICCV} (DCP). DCP uses a Transformer architecture~\cite{Vaswani2017NeurIPS} to compute point-wise features. To limit the influence of a backpropagation cascading effect, i.e. that a difference in gradient in the last layers causes considerably stronger differences in gradients in the early layers, we restrict our focus to the gradients of parameters of the very last layer of the transformer decoder, namely the mean and standard deviation of a Normalization layer.

\PAR{Different Gradients.}
\emph{If Kabsch and our Linear Estimator produce different rotation estimates, then including our layer will produce different gradients}. Without loss of generality, consider the loss function used to train DCP, specifically only the terms that explicitly penalize rotation error. This loss is of the form
\begin{equation}
    \mathcal{L}_{\mat R} = \frac{1}{N_r + 1} \sum_{i=1}^{N_r + 1} \| \mat R_i^\top \mat R_{gt} - \mat I_3\|^2, \label{eq:loss-rotation}
\end{equation}
where $N_r$ is the number of refinements performed with the linear estimator. Consider the simple case where $N_r = 1$ and let us denote Kabsch by $f(\mat P'_t)$ and the Linear Estimator by $g(\mat P'_t, f(\mat P'_t))$, omitting variables in both functions that are not dependent on learnable parameters. The matrix $\mat P'_t$ represents the regressed (target) correspondences by DCP. When Kabsch is used standalone, the gradient will respect the following relationship:
\begin{equation}
    \frac{\partial \mathcal{L}_{\mat R}}{\partial \mat P'_t} = \frac{\partial \mathcal{L}_{\mat R}}{\partial f} \frac{\partial f}{\partial \mat P'_t}.
\end{equation}
Performing a single refinement of the linear estimator, will modify this gradient to
\begin{equation}
    \frac{\partial \mathcal{L}_{\mat R}}{\partial \mat P'_t} = \frac{1}{2}\frac{\partial \mathcal{L}_{\mat R}}{\partial f} \frac{\partial f}{\partial \mat P'_t} + \frac{1}{2}\frac{\partial \mathcal{L}_{\mat R}}{\partial g}\left(\frac{\partial g}{\partial \mat P'_t} + \frac{\partial g}{\partial f}\frac{\partial f}{\partial \mat P'_t}\right).
\end{equation}
In situations where the linear estimator produces estimates similar to Kabsch, we have that $g(\mat P'_t, f(\mat P'_t)) \approx f(\mat P'_t)$, yielding
\begin{align}
    \frac{\partial \mathcal{L}_{\mat R}}{\partial \mat P'_t} &\approx \frac{1}{2}\frac{\partial \mathcal{L}_{\mat R}}{\partial f} \frac{\partial f}{\partial \mat P'_t} + \frac{1}{2}\frac{\partial \mathcal{L}_{\mat R}}{\partial g}\left(0 + \mat I \frac{\partial f}{\partial \mat P'_t}\right) \\
    & \approx \frac{\partial \mathcal{L}_{\mat R}}{\partial f} \frac{\partial f}{\partial \mat P'_t}.
\end{align}
Conversely, we can only have different gradients if the linear estimator produces a different estimate than Kabsch.

\begin{figure}
\begin{center}
\includegraphics[width=0.98\linewidth]{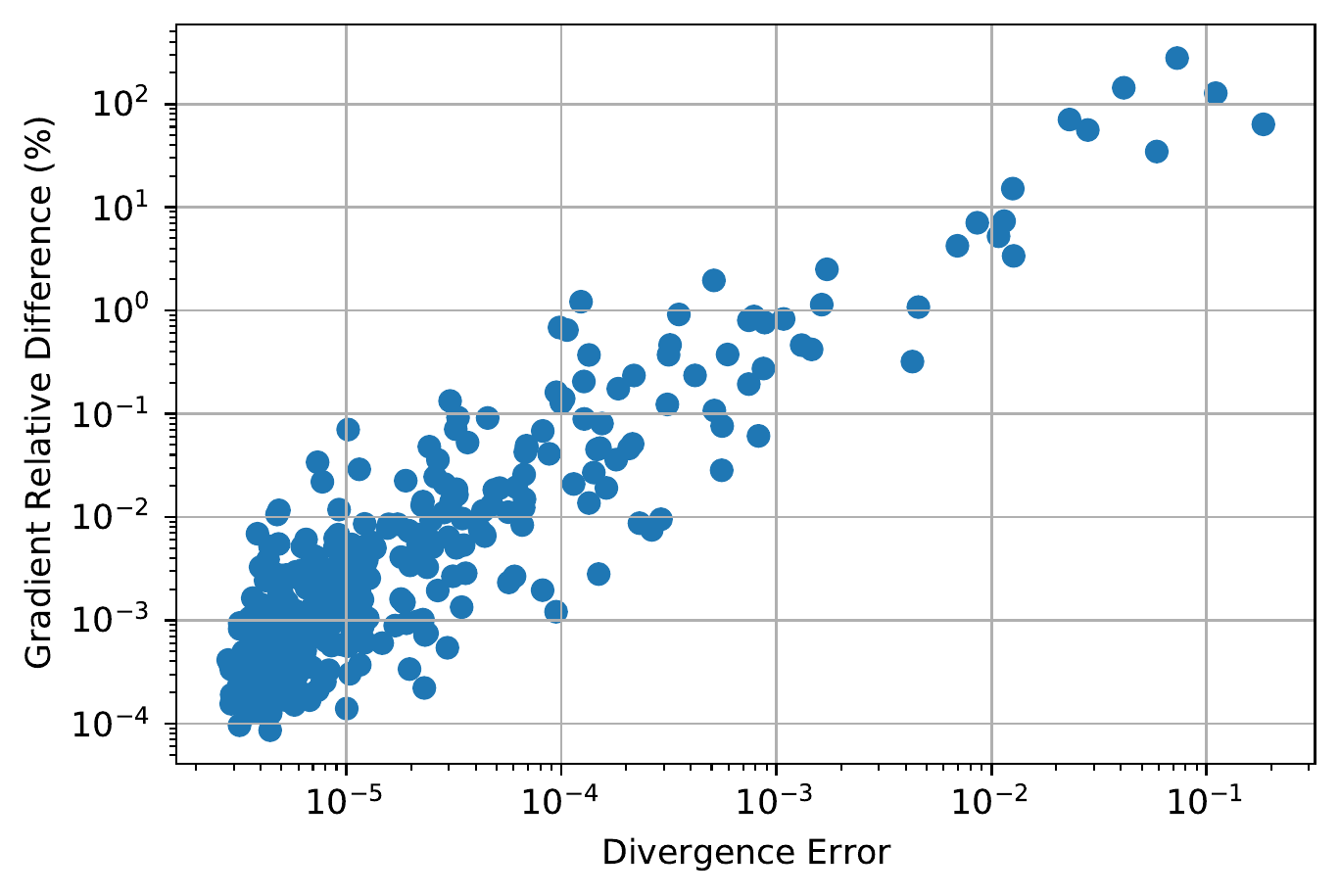}
\end{center}
  \caption{A representation of the gradient relative difference, between employing and not employing our extra layers, as a function of the divergence error between the rotation estimates of the Linear Estimator and Kabsch. This image shows that networks trained with our layers, experience difference gradients whenever the Linear Estimator produces different rotation estimates than Kabsch.}
\label{fig:divergence-corr}
\end{figure}

To confirm this, we conducted an experiment over a single training epoch, where we measured the relative gradient difference as a function of the amount of divergence error. This experiment uses the same training data as in DCP's unseen categories experiment. To measure gradient difference, we perform a forward and backwards pass without our layer, and store Kabsch's gradient. We then do a forward and backward pass with our layers added, storing also this gradient. We proceed with training using our gradient as the descent direction. To report a relative gradient difference, we compute an \emph{element-wise} relative difference between both gradients, concretely
\begin{equation}
    \Delta (\nabla \mathcal{L})_i = \frac{|\nabla \mathcal{L}_{\text{ours}_i} - \nabla \mathcal{L}_{\text{Kabsch}_i}|}{|\nabla \mathcal{L}_{\text{Kabsch}_i}|}.
\end{equation}
We report the mean relative gradient difference, only for the parameters in the last layer. We compute divergence as the chordal distance between Kabsch's and our estimates:
\begin{equation}
    \mathcal{D} = \sum_{i=1}^5 \|\mat R_{\text{ours}_i} - \mat R_\text{Kabsch}\|_F. \label{eq:divergence-error}
\end{equation}
In \autoref{fig:divergence-corr} we can see that as predicted, networks trained with our layers, experience difference gradients whenever the Linear Estimator produces different rotation estimates than Kabsch. In the majority of situations, given a rotation estimate from Kabsch, our estimator will replicate it. However, the linearized constraints make our estimator increasingly sensitive to certain geometric configurations of point clouds. Under these configurations, the estimator will produce a pose estimate that will diverge from Kabsch at each iteration. We stress that in our case, divergence comes paired with the positive effect of facilitating the network to avoid said configurations, something that is also beneficial for Kabsch.

\PAR{Understanding Divergent Cases.}
We have established that divergence from Kabsch is what produces different training outcomes, but we have yet to understand under which conditions our method diverges. At the time of writing, we still do not hold a definitive answer that fully identifies the underlying cause, but we have found certain conditions that establish some empirical upper boundaries on whether divergence has a chance to occur. These mostly depend on the geometric relationship between the unconstrained solution to 
\begin{equation}
    \mat R_u = \argmin_{\mat R} \sum_{i=1}^N w_i \|\tilde{\vec{p_t}}_i  - \mat R \tilde{\vec{p_s}}_i\|^2 \label{eq:problem-rot-unconstrained}
\end{equation}
and the solution produced by Kabsch, that we shall henceforth designate by $\mat R_K \in SO(3)$.

\begin{figure*}
\centering
\begin{minipage}{.5\textwidth}
  \centering
  \includegraphics[width=\linewidth]{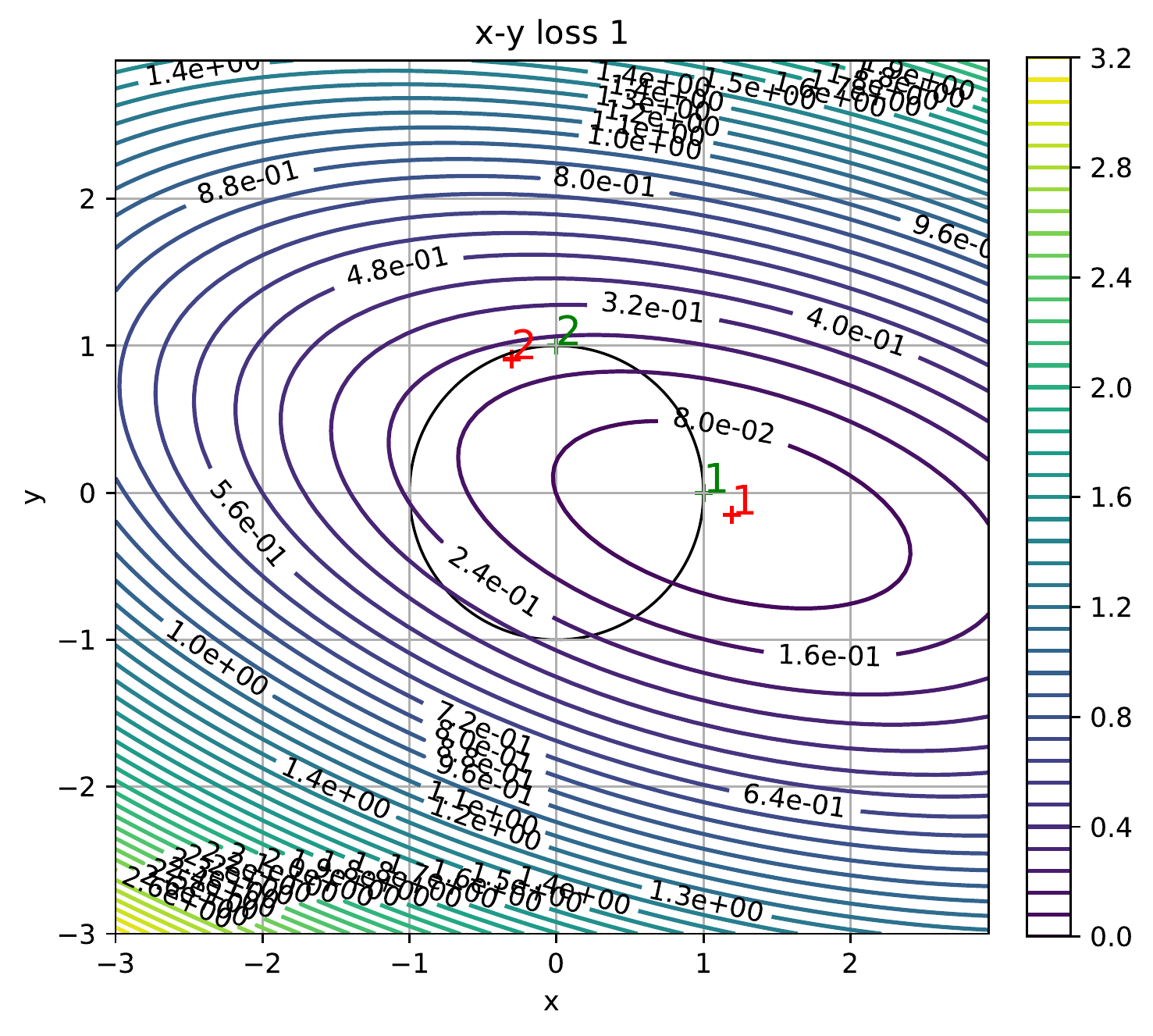}
  \label{fig:loss-landscape-1}
\end{minipage}\hfill%
\begin{minipage}{.5\textwidth}
  \centering
  \includegraphics[width=\linewidth]{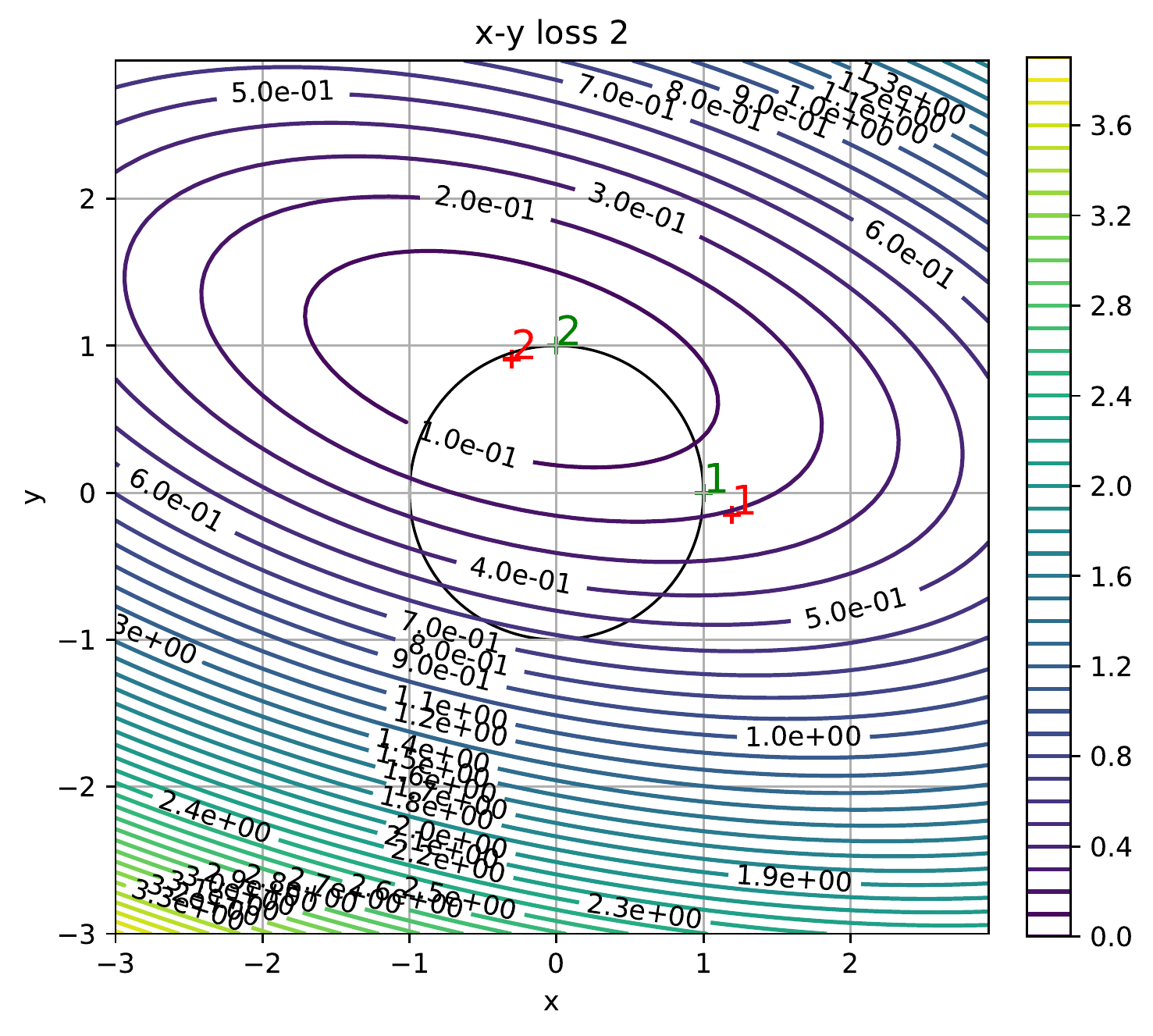}
  \label{fig:loss-landscape-2}
\end{minipage}
\caption{A cross-section of the loss landscape for the first two column vectors of $\mat R$.
\emph{left:} Loss landscape for the column vector $\vec r_1$.
\emph{right:} Loss landscape for the column vector $\vec r_2$.
The figure's axes are aligned with the initialization $\mat R_{t-1}$ provided to the linear estimator. The numbers 1 and 2 indicate the location of each column vector, with red representing the solution of the unconstrained problem $\vec r_{u_i}$ and green the solution from Kabsch $\vec r_{K_i}$. The level set surfaces form quadrics in 3D space. This is a particular example in which the linear estimator does not diverge from the Kabsch's estimate.}
\label{fig:loss-landscape}
\end{figure*}

\begin{figure*}
\centering
\begin{minipage}{.5\textwidth}
  \centering
  \includegraphics[width=\linewidth]{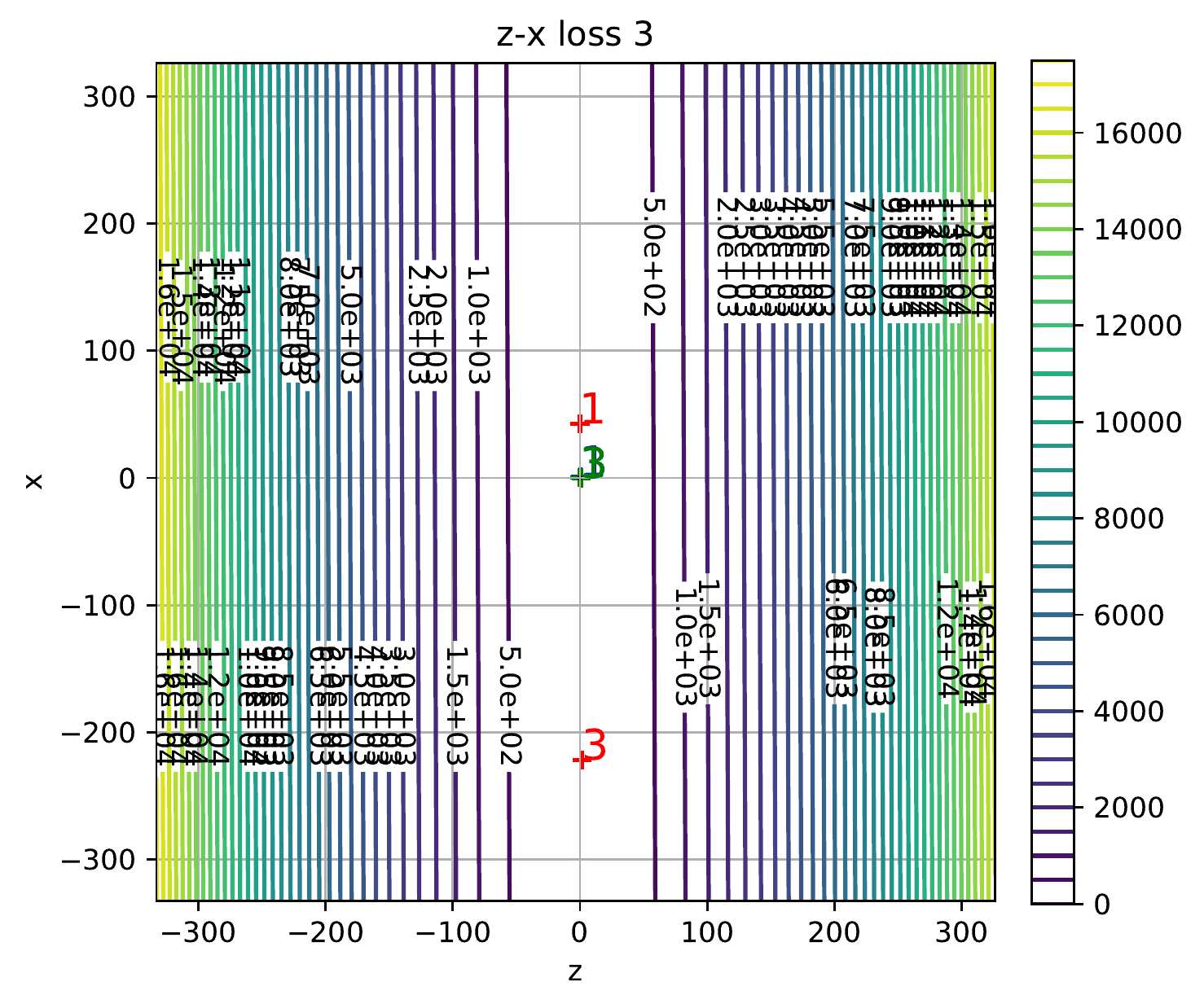}
  \label{fig:divergence-1}
\end{minipage}\hfill%
\begin{minipage}{.5\textwidth}
  \centering
  \includegraphics[width=\linewidth]{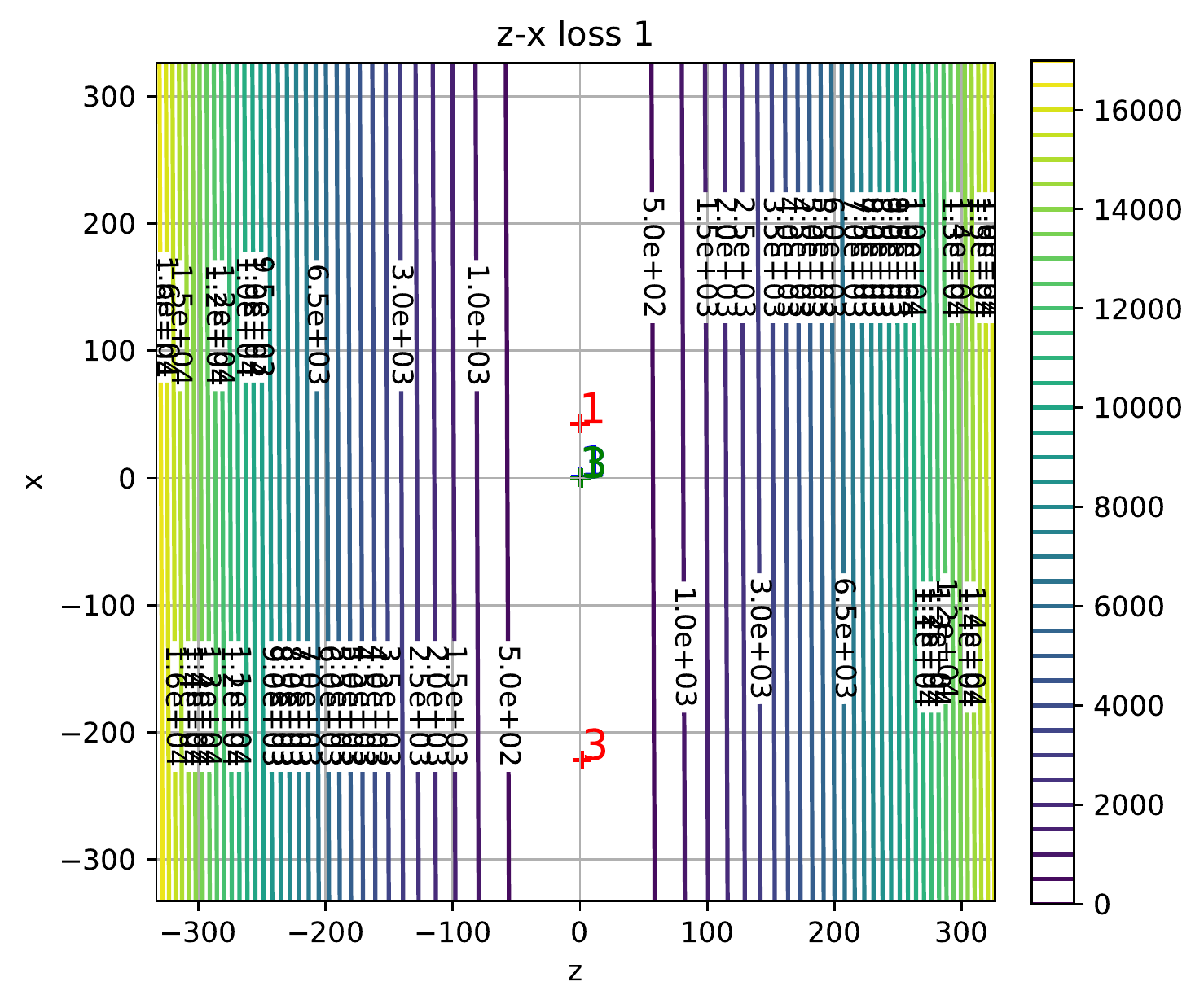}
  \label{fig:divergence-2}
\end{minipage}
\begin{minipage}{.5\textwidth}
  \centering
  \includegraphics[width=\linewidth]{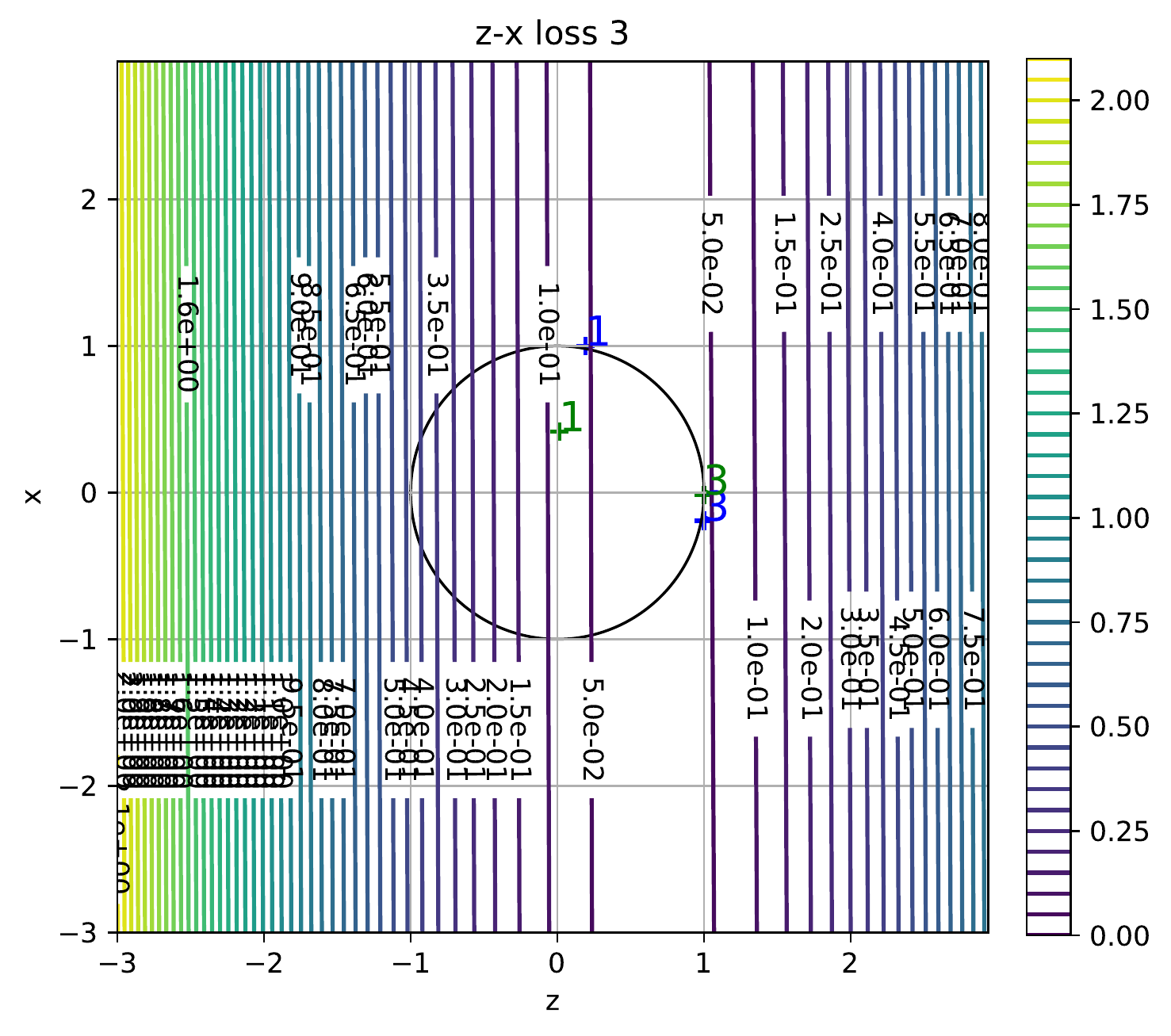}
  \label{fig:divergence-zoomed-1}
\end{minipage}\hfill
\begin{minipage}{.5\textwidth}
  \centering
  \includegraphics[width=\linewidth]{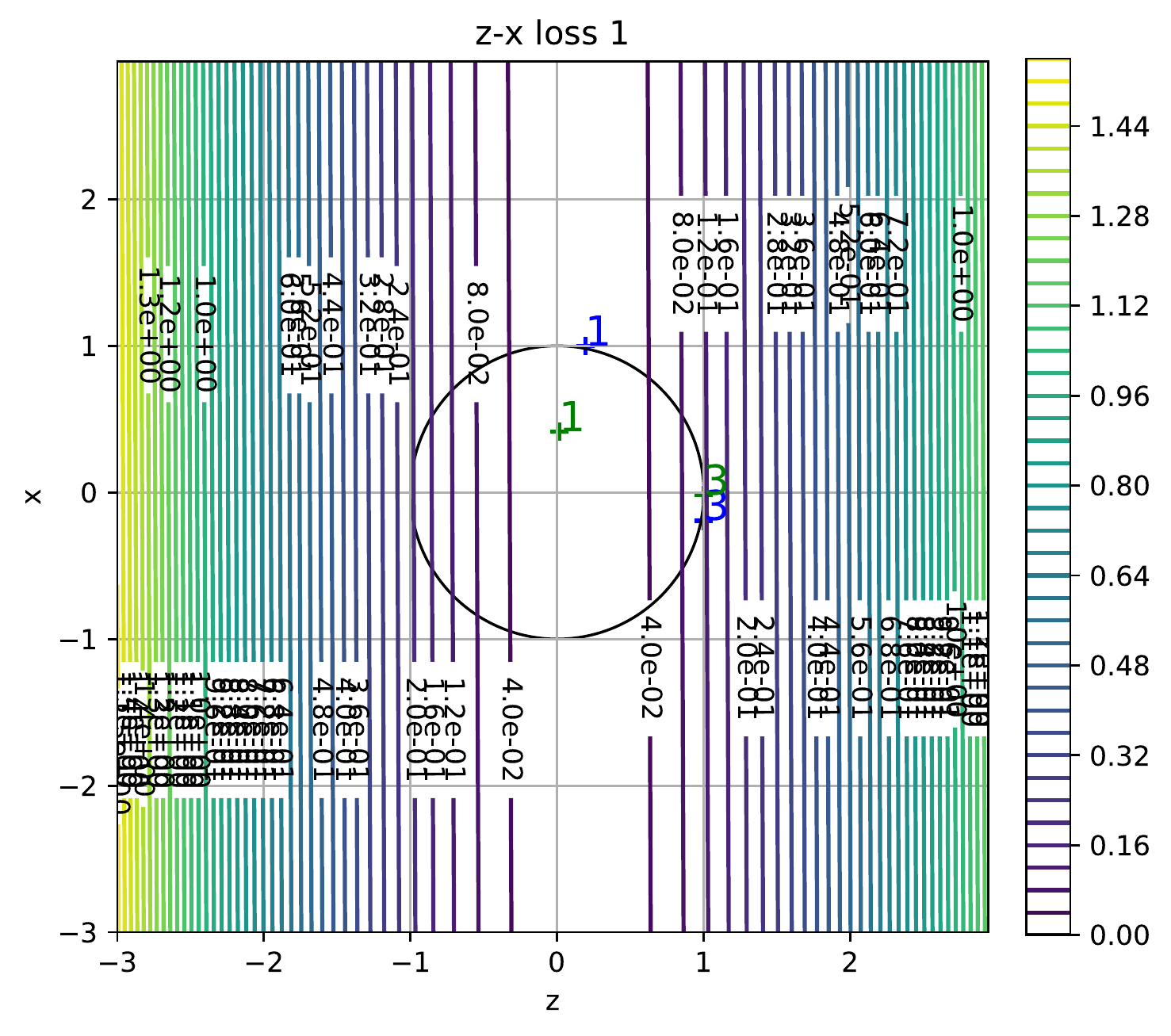}
  \label{fig:divergence-zoomed-2}
\end{minipage}
\caption{A cross-section of the loss landscape for the first and third column vectors of $\mat R$. 
\emph{top-left:} Loss landscape for the $\vec r_3$.
\emph{top-right:} Loss landscape for the $\vec r_1$.
\emph{bottom-left:} Loss landscape for the $\vec r_3$. A zoomed-in view of top-left.
\emph{bottom-right:} Loss landscape for the $\vec r_1$. A zoomed-in view of top-right.
The numbers 1 and 3 indicate the location of each column vector, with red representing the solution of the unconstrained problem $\vec r_{u_i}$, blue representing the solution of the linear estimator $\vec r_{i}$ and green the solution from Kabsch $\vec r_{K_i}$. The figure's axes are aligned with the initialization $\mat R_{t-1}$ provided to the linear estimator, and explain why the solution from Kabsch is no longer contained inside the unit circle. In this particular example, the linear estimator diverges from Kabsch's estimate. Contrary to \autoref{fig:loss-landscape}, note how the unconstrained solution is considerably distant from Kabsch's estimate and how the level set curves appear to be almost parallel, usually a sign that the correspondences in the target point cloud are close to being distributed along a linear subspace in 3D space, e.g. a plane or a line. In this image we can also confirm that the solutions from the linear estimator respect the constraints shown in \autoref{fig:feasible-set}.}
\label{fig:divergence}
\end{figure*}

It is important to recognize that the optimization problem in Eq.~\eqref{eq:problem-rot-unconstrained} is solving three independent optimization problems. To illustrate that, we employ an alternative formulation of the problem in Eq.~\eqref{eq:problem-rot-unconstrained}
\begin{align}
    \mat R_u &= \argmin_{\mat R} \sum_{i=1}^N w_i \|\mat R^\top \tilde{\vec{p_t}}_i  - \tilde{\vec{p_s}}_i\|^2 \label{eq:problem-rot-unconstrained-2} \\
    &= {\underbrace{\left(\sum_{i=1}^N w_i \tilde{\vec{p_t}}_i \tilde{\vec{p_t}}_i^\top\right)}_{\mat G}}^{-1} \underbrace{\left(\sum_{i=1}^N w_i \tilde{\vec{p_t}}_i \tilde{\vec{p_s}}_i^\top\right)}_{\mat F},
\end{align}
with matrices $\mat R_u, \mat F \in \R^{3\times 3}$ and $\mat G \in \mathbb{S}^3$. Each column of matrix $\mat R$ will independently  pick the best location in 3D that will minimize its correspondence error. Each column $\vec r_{u_i} \in \R^3$ of matrix $\mat R_u$ is given by
\begin{equation}
    \vec r_{u_i} = \mat G^{-1} \vec f_i
\end{equation}
where $\vec f_i \in \R^3$ is the i-th column of $\mat F$. So each column $\vec r_{u_i}$ has its own distinct loss landscape in $\R^3$ that can be expressed as
\begin{equation}
    \mathcal{L}_{\vec r_{u_j}} = \sum_{i=1}^N w_i (\tilde{\vec{p_t}}_i^\top \vec r_j -  \tilde{\vec{p_s}}_{ij})^2 \quad \text{with } 1 \leq j \leq 3.
\end{equation}
Note the usage of index $j$ to represent the column index, to avoid a clash with index $i$ that denotes correspondences. The level set surfaces of these loss landscapes form quadrics in 3D space as exemplified in \autoref{fig:loss-landscape}.

\autoref{fig:loss-landscape} shows an example where the linear estimator produces an estimate that does not diverge from Kabsch's estimate. In contrast, in \autoref{fig:divergence} we show a particular example where the linear estimator diverged considerably. This example captures some of the representative aspects that cause the linear estimator to diverge: the unconstrained solutions is considerably far away from Kabsch's estimate, usually a consequence of level set quadrics that are close to being degenerate due to the target point cloud not spanning full 3D space; some of the unconstrained solution column vectors are close to being orthogonal to their corresponding column vectors in Kabsch's estimate. We explore these two cases further.

\begin{figure}
\begin{center}
\includegraphics[width=0.98\linewidth]{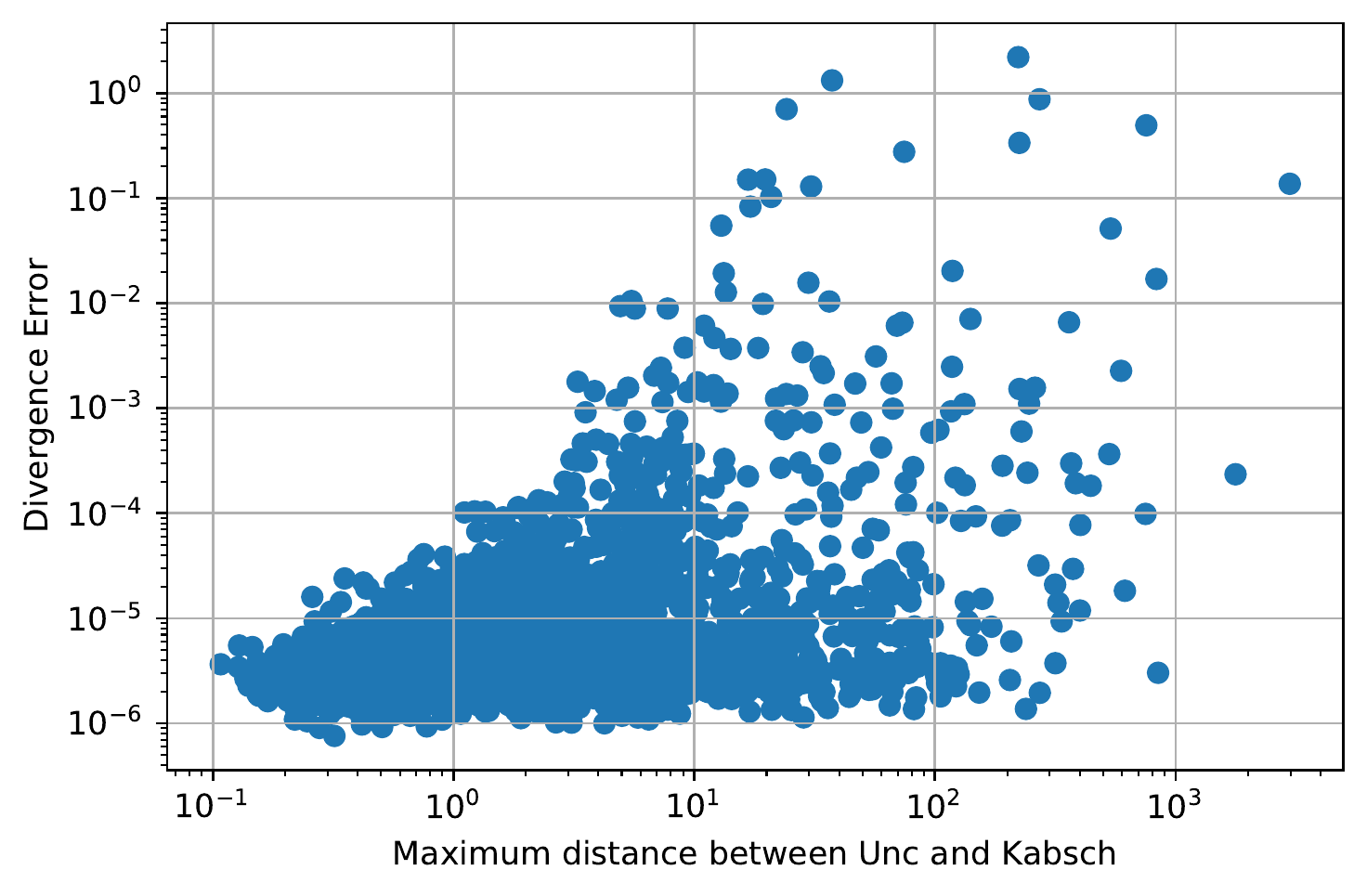}
\end{center}
  \caption{A representation of the divergence error,  as a function of the maximum distance across all column vectors, between Kabsch's and the unconstrained solution. The distance establishes a practical upper bound on the divergence error, that increases when the distance between both solutions also increases.}
\label{fig:distance-corr}
\end{figure}
\begin{figure}
\begin{center}
\includegraphics[width=0.98\linewidth]{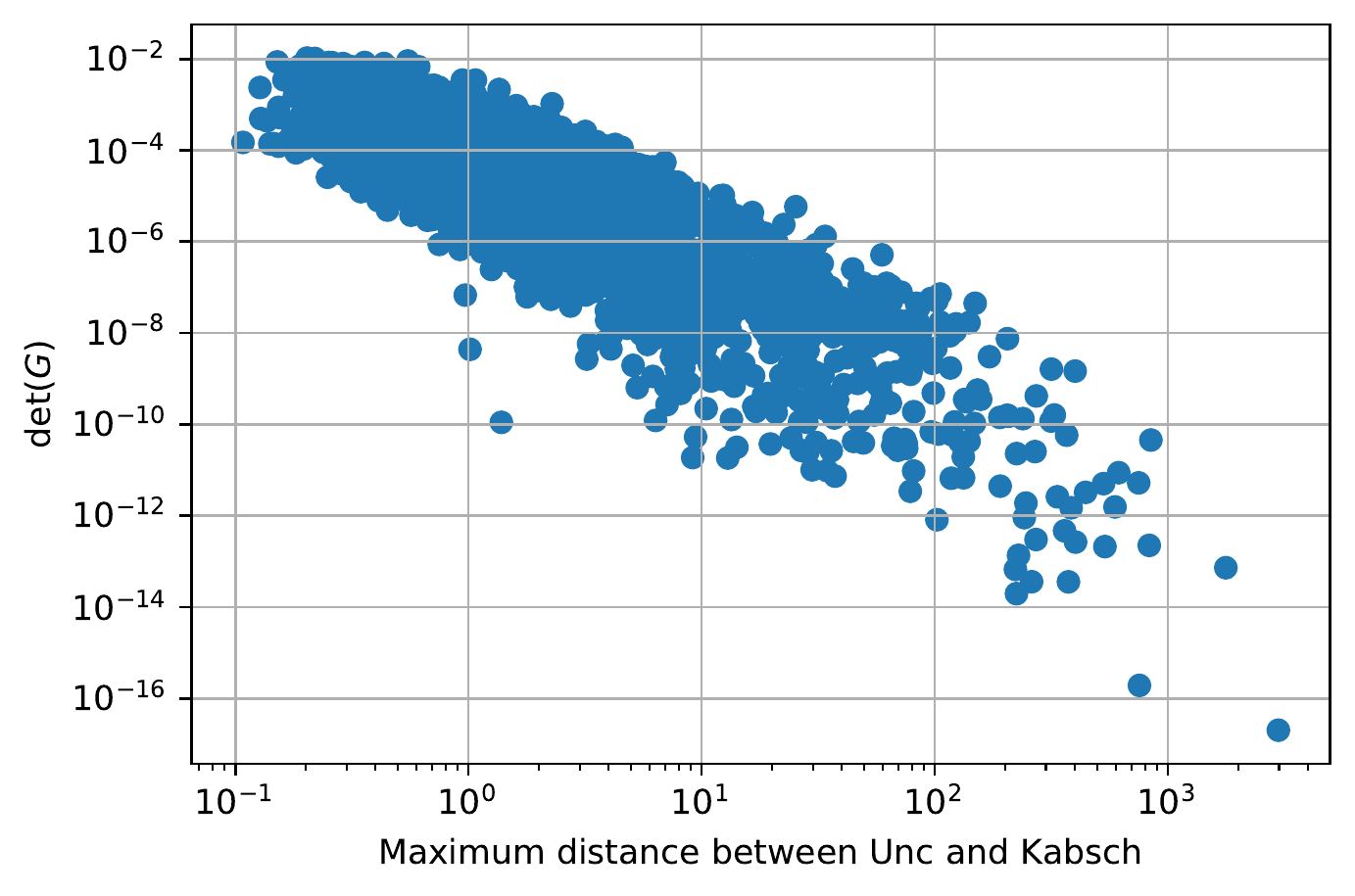}
\end{center}
  \caption{A representation of $\det(\mat G)$, as a function of the maximum distance across all column vectors, between Kabsch's and the unconstrained solutions. The closer the points in the target point cloud are to span only a linear subspace of 3D, like a plane or a line, the closer the $\det(\mat G)$ will be to the value 0. When that happens, our the level set quadrics approximate degeneracy and the unconstrained solution shifts considerably in space.}
\label{fig:distance-det-corr}
\end{figure}

\PAR{Distance between the Unconstrained and Kabsch's Solutions.}
We conducted an experiment over a single training epoch, where we measured the maximum distance between Kabsch's and the unconstrained solutions, across all three column vectors. This experiment uses the same training data as in DCP's unseen categories experiment. We map the divergence error from Eq.~\eqref{eq:divergence-error} to the maximum distance between both solutions, computed as
\begin{equation}
    d_{\text{unc, Kabsch}} = \max_i \| \vec r_{u_i} - \vec r_{K_i}\|,
\end{equation}
and show it in \autoref{fig:distance-corr}. We can see that this distance establishes a practical upper bound on the divergence error, that increases when the distance between both solutions also increases. In \autoref{fig:distance-det-corr}, we validate our claim of the strong correlation between the unconstrained solution distance and how this usually manifests itself when the level set quadrics are close to being degenerate.

\begin{figure}
\begin{center}
\includegraphics[width=0.98\linewidth]{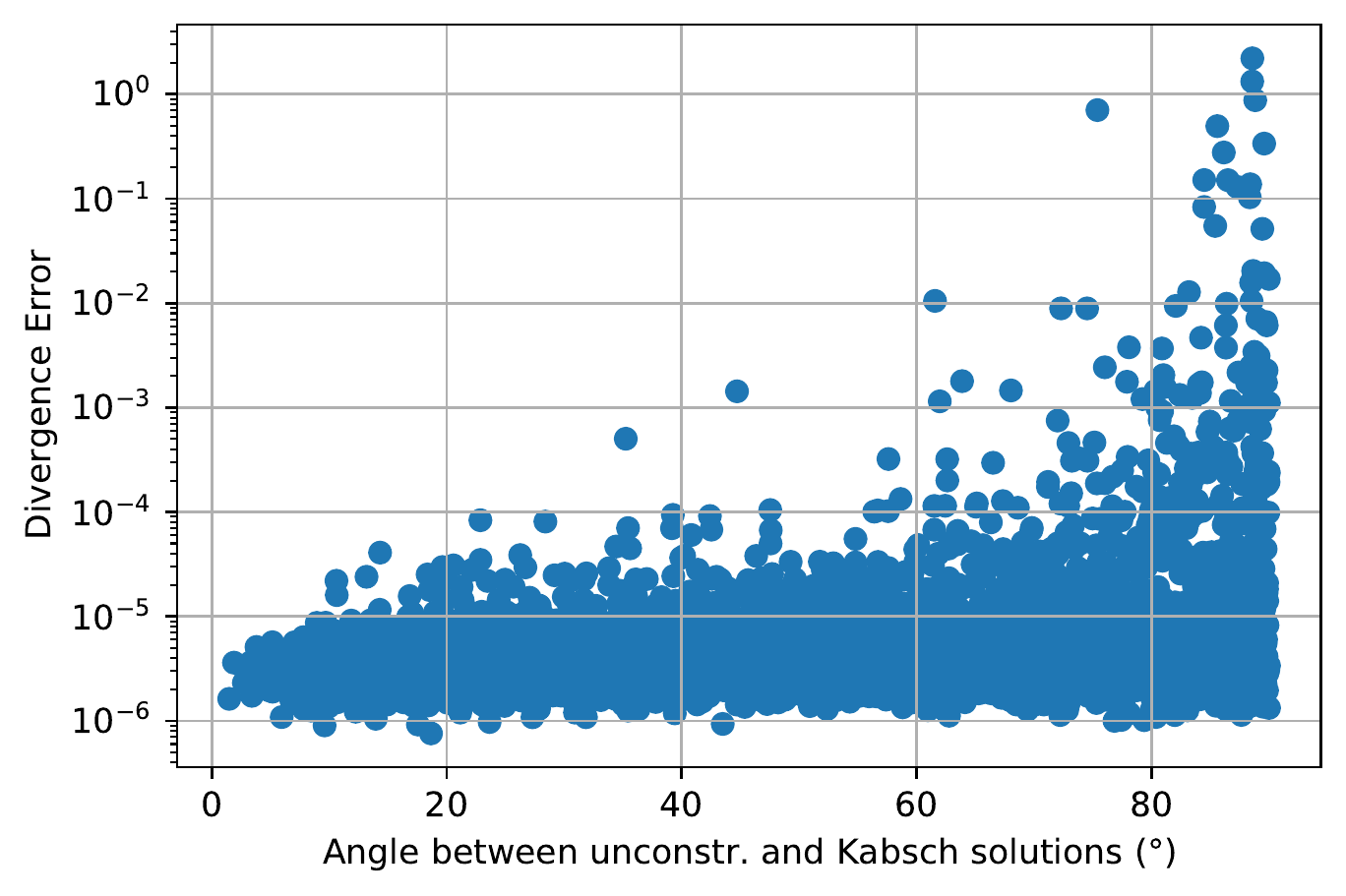}
\end{center}
  \caption{A representation of the divergence error, as a function of the maximum angle across all column vectors, between Kabsch's and the unconstrained solution. Similar to \autoref{fig:distance-corr}, the angle also establishes a practical upper bound on the divergence error, that increases when the angle between both solutions also increases.}
\label{fig:angle-corr}
\end{figure}

\PAR{Angle between the Unconstrained and Kabsch's Solutions.}
We conducted an experiment over a single training epoch, where we measured the maximum angle between Kabsch's and the unconstrained solutions, across all three column vectors. This experiment also relies on the same training data as in DCP's unseen categories experiment. We map the divergence error from Eq.~\eqref{eq:divergence-error} to the maximum angle between both solutions, computed as
\begin{equation}
    \angle_{\text{unc, Kabsch}} = \max_i \frac{180}{\pi}\arccos \left(\left|\frac{\vec r_{u_i}}{\|\vec r_{u_i}\|}^\top \frac{\vec r_{K_i}}{\|\vec r_{K_i}\|} \right|\right),
\end{equation}
and show it in \autoref{fig:angle-corr}. We can see the angle also establishes a practical upper bound on the divergence error, that increases when the angle between both solutions also increases.

\section{Additional Experiments}
\label{sec:experiments}

\subsection{Deep Closest Point with ModelNet40 Data}

\noindent{\bf Alignment for identical point clouds.} In \autoref{table:full:modelnet40}, we report the point cloud registration results obtained on the instances of CAD models that were held-out during the model training. 
In this setting, we are aligning identical point clouds; therefore, perfect correspondences between them exist. As can be seen, with the addition of our refinements, we obtain a performance that is on-par with the original method. We marginally improve the rotation error by 0.07\% at the marginal cost of translation accuracy of 0.19\%, when normalized by the maximum magnitude of the rotation and translations sampled. 
When the problems is simple for the underlying network, our method does not provide any assistance, but it will not make the results worse. This allows to blindly use it as an add-on that can only improve performance.
In the experiments presented in the main paper, we show that in more challenging scenarios, our layer provides more meaningful improvements.

\begin{table}[t] 
\begin{center}
\resizebox{\linewidth}{!}{%
    \begin{tabular}{lccccr}
    \toprule
    Model & RMSE($\mat R$)\degree & MAE($\mat R$)\degree  & RMSE($\vec t$) & MAE($\vec t$)\\ 
    \midrule
    ICP &  29.914835 & 23.544817 & 0.290935 & 0.248755 \\
    Go-ICP \cite{Yang15TPAMI} & 11.852313 & 2.588463 & 0.025665 & 0.007092 \\
    FGR \cite{Zhou16ECCV} &  9.362772 & 1.999290 & 0.013939 & 0.002839 \\
    PointNetLK \cite{Goforth19CVPR} &   15.095374 &  4.225304 & 0.022065 & 0.005404 \\
    \midrule
    DCP-v2 & 1.093971 & \textbf{0.751517} & \textbf{0.001717} & \textbf{0.001173} \\
    \textbf{DCP-v2 + ours}  & \textbf{1.063893} & 0.760524 & 0.002677 & 0.001865 \\
    \bottomrule
    \end{tabular}%
}
\end{center}
\caption{Deep Closest Point on ModelNet40:  Test on unseen point clouds with perfect correspondences. We marginally improve the rotation error by 0.07\% at the marginal cost of translation accuracy of 0.19\%, when normalized by the maximum magnitude of the rotation and translations sampled. The problem is too simple for our layer to provide meaningful improvements to the baseline, but it will not make the results worse, allowing it to blindly used as an add-on that can only improve match or improve performance. \label{table:full:modelnet40}}
\end{table}

\subsection{RPM-Net with ModelNet40 Data}

\noindent{\bf Alignment under Gaussian noise.} We sample a different set of 1024 points (from the original 2048) for each point cloud
and add Gaussian noise $\mathcal{N}(0, 0.01^2)$ independently to both, clamped at $[-0.05, 0.05]$. 
\begin{table}
\small
\begin{center}
\setlength\tabcolsep{4.5pt}
\begin{tabularx}{\linewidth}{X | c c c c c}
  \toprule
  Method & \multicolumn{2}{c}{Anisotropic err.} & \multicolumn{2}{c}{Isotropic err.} & $\tilde{CD}$\\
   & (Rot.) & (Trans.) & (Rot.) & (Trans.) &  \\
  \midrule
  ICP         &  3.414  &  0.0242  &  6.999  &  0.0514  &  0.00308  \\
  RPM         &  1.441  &  0.0094  &  2.994  &  0.0202  &  0.00083  \\
  FGR         &  1.724  &  0.0120  &  2.991  &  0.0252  &  0.00130  \\
  PointNetLK  &  1.528  &  0.0128  &  2.926  &  0.0262  &  0.00128  \\
  DCP-v2      &  4.528  &  0.0345  &  8.922  &  0.0707  &  0.00420  \\
  \midrule
  RPM-Net         & 0.343      & \textbf{0.0030}       & \textbf{0.664}      & \textbf{0.0062}       &   \textbf{0.00063} \\
  RPM-Net + Ours  & \textbf{0.342}      & \textbf{0.0030}       & \textbf{0.664}      & \textbf{0.0062}       &   \textbf{0.00063} \\
  \bottomrule
\end{tabularx}
\end{center}
\caption{RPM-Net on ModelNet40: Performance on data with Gaussian noise. The Chamfer distance using groundtruth transformations is 0.00055. 
Our network provides no improvement in this case because the problem is simple for the matching network, but it also does not hinder performance.
}
\label{table:rpm-noisy-performance}
\end{table}
We show the results in \autoref{table:rpm-noisy-performance}.
In this experiment, we do not provide any measurable improvement to the baseline method. Similar to DCP, this happens when the correspondence problem is too simple. However, the results encourage the idea that we do not incur a penalty in including our layer and that it can be blindly applied as an add-on. We also evaluated increasing the magnitude of Gaussian error at test time, but both approaches produce similar results.


\subsection{Ablation Studies}
\label{sec:ablations}

Our method is governed by two critical design choices: how many refinement iterations should we conduct and whether to impose a loss in all poses output by our method or only in the last one. In this section we show how both these decisions affect registration performance.
We evaluate the choice of performing 1, 2, 5 and 10 refinement iterations. We carry-over the experimental setup from DCP with unseen categories and from RPM-Net on partially visible data with noise. We report RMSE for both rotation and translation for these variants. We report results for DCP in \autoref{table:exp-dcp-ablation} and for RPM-Net in \autoref{table:rpm-crop-ablation}. In both cases, applying the pose loss to all poses produced by the network, in conjunction with performing 5 iterations of our method produces the best pose error.

\begin{table}[t] 
\footnotesize
\begin{center}
\begin{tabular}{lccc}
\toprule
Iters. / Loss & $\Delta \vec{p}$ & $\Delta \mat R_{\text{ani}}$ & $\Delta \vec t$\\ 
\midrule
1 / all & 0.491894 & 5.472425 & 0.005496\\
2 / all & 0.491909 & 5.573030 & 0.007119\\
5 / all & \textbf{0.491878} & \textbf{2.051718} & \textbf{0.004543}\\
10 / all & 0.492074 & 5.558087 & 0.014462\\
\midrule
1 / last & 0.491913 & 3.791935 & 0.006351\\
2 / last & 0.491890 & 2.485598 & 0.004585\\
5 / last & 0.491963 & 4.859206 & 0.009861\\
10 / last & 0.492326 & 6.622592 & 0.021234\\
\bottomrule
\end{tabular}
\end{center}
\caption{Ablations on the Deep Closest Point unseen categories experiment. We evaluate the influence of the number of refinement iterations used, as well as the effect of employing a loss term to all or just the last pose produced by the combination of Kabsch and our method.
We present results for the mean point distance $\Delta \vec p$,
and RMSE for rotation $\Delta \mat R_{\text{ani}}$ 
and translation $\Delta \vec t$ 
($\mathcal{L}_2$ norm) errors.
\label{table:exp-dcp-ablation}}
\vspace{-1.5em}
\end{table}

\begin{table}
\small
\begin{center}
\setlength\tabcolsep{4.5pt}
\begin{tabularx}{\linewidth}{X | c c c c c}
  \toprule
  Method & \multicolumn{2}{c}{Anisotropic err.} & \multicolumn{2}{c}{Isotropic err.} & $\tilde{CD}$\\
   & (Rot.) & (Trans.) & (Rot.) & (Trans.) &  \\
  \midrule
    1 / all & 0.8944    & 0.00877    & 1.704    & 0.0184    & 0.00089 \\
    2 / all & 0.8851    & 0.00861    & 1.686    & 0.0183    & 0.00091 \\
    5 / all & \textbf{0.8318}    & \textbf{0.00805}    & \textbf{1.577}    & \textbf{0.0169}    & \textbf{0.00085} \\
    10 / all & 0.8473    & 0.00815   & 1.604    & 0.0172    & \textbf{0.00085} \\
  \midrule
    1 / last & 0.8798    & 0.00833    & 1.661    & 0.0175   & 0.00087 \\
    2 / last & 0.8792    & 0.00835    & 1.695    & 0.0176   & \textbf{0.00085} \\
    5 / last & 0.8570    & 0.00843    & 1.634    & 0.0177   & 0.00087 \\
    10 / last & 0.8876    & 0.00839    & 1.687    & 0.0177   & 0.00087 \\
\bottomrule
\end{tabularx}
\end{center}
\caption{Ablation RPM-Net on ModelNet40: Performance on partially visible data with noise. The Chamfer distance using groundtruth transformations is 0.00055.}
\label{table:rpm-crop-ablation}
\end{table}

\section{Measuring Correspondence Improvement}
\label{sec:correspondences}

\begin{figure}
\begin{center}
\includegraphics[width=\linewidth]{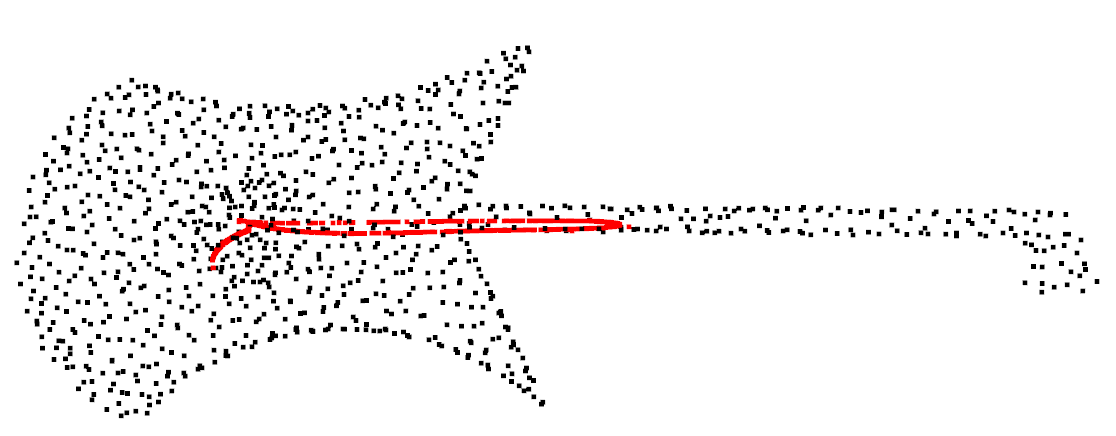}
\end{center}
  \caption{A qualitative example of the how correspondences are generated by Deep Closest Point. In the image we see point clouds of two different colors: black and red. We cherry-pick an example with the lowest pose estimation error, $\angle \Delta \mat R_\text{iso} = 0.2712\degree$, $\Delta \vec{t} = 0.0002$. \textbf{Black}: Point cloud generated by applying the ground-truth transformation to the source point cloud i.e., each point is given by $\vec{p}_{t_i} = \mat R_\text{gt} \vec{p}_{s_i} + \vec{t}_\text{gt}$. \textbf{Red}: The correspondences produced by the network to perform the registration task, where each point represents $\vec{p}'_{t_i}$.}
\label{fig:correspondences_s}
\vspace{-1.5em}
\end{figure}

In the main paper, we made the claim that evaluating correspondence quality simply based on the Euclidean distance can be misleading. This is because improvements in this metric do not necessarily translate in improvements in pose. In fact, it is possible to engineer a particular case that for a higher average Euclidean distance error, the network produces a better pose estimate. In light of this, we make the argument that only pose error can accurately represent a measure of correspondence quality improvement, for the task of point cloud registration.  To establish an initial intuition behind our claims we refer to \autoref{fig:correspondences_s}, also present in the main paper. In here, we show a cherry picked example where the pose error is particularly small. Contrary to intuition, the regressed target points (red) hardly resemble the ground truth target points (black) and yet the network is still able to estimate an almost perfect pose. This confirms that a seemingly high positional error between regressed and ground truth target points does not imply a bad pose estimate. In fact, \autoref{fig:correspondences_s} suggests that in order to retrieve an accurate pose estimate, it only matters that the centroids and principal directions of both point sets are relatively consistent.

Recall that for each point $\vec{p}_{s_i}$ in the source point cloud, both DCP and RPM-Net regress the coordinates of its corresponding point, expressing it as $\vec{p}'_{t_i} = \sum_{j = 1}^N \alpha_{ij} \vec{p}_{t_j}$, where $\alpha_{ij}$ is the probability of point $\vec{p}_{s_i}$ matching $\vec{p}_{t_i}$. The mean subtracted version of these pairs of correspondences $\tilde{\vec{p}}_{s_i}$ and $\tilde{\vec{p}}'_{t_i}$ are used as input to Kabsch. The Kabsch algorithm~\cite{gower75Psychometrika} provides a closed-form to the problem in Eq.~\eqref{eq:simplified_s}. Given correspondences, Kabsch computes a globally optimal solution via SVD, as follows:
\begin{align}
    \mat H &= \sum_{i = 1}^N w_i\tilde{\vec{p'_t}}_i \tilde{\vec{p_s}}^\top_i  \label{eq:kabsch-input_s}\\
    \mat U, \mat S, \mat V &= \operatorname{svd}(\mat H) \\
    \mat R &= \mat U \operatorname{diag} ([1, 1, \det(\mat U \mat V^\top)])\mat V^\top.
\end{align}
The operator $\operatorname{diag(\:)}$ produces a diagonal square matrix, in which the input vector represents the diagonal.
To produce a correct rotation estimate, it is not necessary that $\forall i : \|\tilde{\vec{p}}'_{t_i} - \mat R \tilde{\vec{p}}_{s_i}\| = 0$. Furthermore, Kabsch is a method that is invariant to scale. Multiplying the source and target point clouds by arbitrary non-negative scalars will produce the same rotation matrix, because these positive scalars will be ``absorbed" by the diagonal matrix of singular values $\mat S$.

To further stress this idea, consider a problem composed of (already centered) point clouds $\tilde{\mat P}'_t, \tilde{\mat P}_s \in \R^{N\times3}$, with each row $\tilde{\vec p}'_{t_i}, \tilde{\vec p}_{s_i} \in \R^3$ representing a corresponding point, for which we already have extracted the optimal rotation $\mat R$ using Kabsch. Let us define the mean squared correspondence error as
\begin{equation}
    d_0 = \frac{1}{N} \sum_{i = 1}^N \|\tilde{\vec p}'_{t_i} - \mat R \tilde{\vec p}_{s_i}\|^2.
\end{equation}
From the previous paragraph, we know that if we multiply $\tilde{\mat P}'_t$ by $a \in \R_+$, the optimal rotation that minimizes the correspondence error remains unchanged. We are now interested in finding the mean squared correspondence error with this new scaled point cloud.
\begin{align}
    d_a &=  \frac{1}{N} \sum_{i = 1}^N \|a \tilde{\vec p}'_{t_i} - \mat R \tilde{\vec p}_{s_i}\|^2 \\
    &= \frac{1}{N} \sum_{i = 1}^N a^2 \tilde{\vec p}_{t_i}^{'\top} \tilde{\vec p}'_{t_i} - 2a\tilde{\vec p}_{t_i}^{'\top} \mat R \tilde{\vec p}_{s_i} + \tilde{\vec p}_{s_i}^\top \mat R^\top \mat R \tilde{\vec p}_{s_i} \\
    &= d_0 + \frac{1}{N} \sum_{i = 1}^N (a^2 - 1)\tilde{\vec p}_{t_i}^{'\top} \tilde{\vec p}'_{t_i} - 2(a - 1) \tilde{\vec p}_{t_i}^{'\top} \mat R \tilde{\vec p}_{s_i} \\
    &= d_0 + \frac{a^2 - 1}{N} \sum_{i = 1}^N \tilde{\vec p}_{t_i}^{'\top} \tilde{\vec p}'_{t_i} - \frac{2}{a + 1} \tilde{\vec p}_{t_i}^{'\top} \mat R \tilde{\vec p}_{s_i} \\
    &= d_0 + \underbrace{\frac{a^2 - 1}{N} \sum_{i = 1}^N \tilde{\vec p}_{t_i}^{'\top} \left(\tilde{\vec p}'_{t_i} - \frac{2}{a + 1} \mat R \tilde{\vec p}_{s_i}\right)}_{\Delta d}. \label{eq:scaled-correspondence-distance}
\end{align}
From Eq.~\eqref{eq:scaled-correspondence-distance}, one can see that as long as all points $\tilde{\vec p}_{s_i}$ are finite, we will always be able find a large enough $a$ that ensures $\Delta d > 0$. In fact, we can make $\Delta d$ arbitrarily large, effectively increasing the mean squared correspondence error by an arbitrary amount, without incurring in any additional pose error. 

Despite the arguments presented, in the interest of completeness, we evaluate the quality of correspondences based on the average point distance, when DCP is trained with and without our layers. 
We revisit the DCP's Gaussian noise experiment, where noise is added independently to one of the point clouds at test time.
\begin{table}
\small
\begin{center}
\begin{tabular}{l c c}
  \toprule
  Metric & DCP & DCP + Ours\\
  \midrule
  Abs. Rot. (\degree) &  10.565127 & \textbf{8.030258}\\
  Rel. Rot. (\%) & 27.130154 & \textbf{20.763237}\\
  \midrule
  Abs. Trans. & \textbf{0.005020} & 0.005533\\
  Rel. Trans. (\%) & \textbf{1.188571} & 1.303502\\
  \midrule
  Abs. Corr. & 0.522025 & \textbf{0.515820}\\
  Rel. Corr. (\%) & 96.776123 & \textbf{95.725166}\\
  \bottomrule
\end{tabular}
\end{center}
\caption{Mean absolute and relative rotation (Rot.), translation (Trans.) and correspondence position (Corr.) errors, averaged over all data samples in ModelNet40's testing set. DCP represents the network trained with its original architecture using the pretrained model supplied the authors of the paper. DCP + Ours represents a network trained with our proposed layer after the Kabsch. We show that a  1\% improvement in correspondence error results in a 7\% improvement in rotation error.} 
\label{table:exp-correspondence-improvements}
\end{table}
We present results for mean correspondence position, isotropic rotation, and translation errors in \autoref{table:exp-correspondence-improvements}. The relative errors  are normalized \wrt ground-truth values. Despite the seemingly marginal improvement in correspondence error, this produces a significant improvement in the quality of pose estimated. We improve the rotation error by 7\% just from a 1\% improvement in correspondence error.